\documentclass[letterpaper, 10 pt, conference]{support/ieeeconf}
\usepackage{times}
\usepackage[pdftex]{graphicx}
\usepackage{subfigure}
\usepackage{amsmath,amssymb,amsopn,amstext,amsfonts}
\usepackage{cancel}
\usepackage[space]{cite}
\usepackage{pdfsync}
\usepackage{balance}
\usepackage{color}
\usepackage{mathtools}
\usepackage{algpseudocode}
\usepackage{algorithm} 

\algnewcommand\algorithmicforeach{\textbf{for each}}
\algdef{S}[FOR]{ForEach}[1]{\algorithmicforeach\ #1\ \algorithmicdo}

\usepackage{bm}

\usepackage{diagbox}
\usepackage{float}
\usepackage{epstopdf}
\usepackage{pifont}
\usepackage{fixltx2e}
\usepackage{mathrsfs}
\usepackage{multirow}
\usepackage{url}
\usepackage{verbatim}
\usepackage[linkcolor=black,citecolor=black,urlcolor=black,colorlinks=true]{hyperref}

\graphicspath{{figures/}}
\DeclareGraphicsExtensions{.png,.jpg,.eps,.pdf}
\IEEEoverridecommandlockouts
\title{\LARGE \bf Neither Fast Nor Slow: How to Fly Through Narrow Tunnels}
\author{Luqi Wang, Hao Xu, Yichen Zhang and Shaojie Shen
\thanks{All authors are with the Department of Electronic and Computer Engineering, Hong Kong University of Science and Technology, Hong Kong, China. {\tt\footnotesize $\{$lwangax, hxubc, yzhangec$\}$@connect.ust.hk, eeshaojie@ust.hk}.}%
}
\begin{document}

\maketitle
\thispagestyle{empty}
\pagestyle{empty}

\begin{abstract}
Nowadays, multirotors are playing important roles in abundant types of missions. During these missions, entering confined and narrow tunnels that are barely accessible to humans is desirable yet extremely challenging for multirotors. The restricted space and significant ego airflow disturbances induce control issues at both fast and slow flight speeds, meanwhile bringing about problems in state estimation and perception. Thus, a smooth trajectory at a proper speed is necessary for safe tunnel flights. To address these challenges, in this letter, a complete autonomous aerial system that can fly smoothly through tunnels with dimensions narrow to 0.6 m is presented. The system contains a motion planner that generates smooth mini-jerk trajectories along the tunnel center lines, which are extracted according to the map and Euclidean Distance Field (EDF), and its practical speed range is obtained through computational fluid dynamics (CFD) and flight data analyses. Extensive flight experiments on the quadrotor are conducted inside multiple narrow tunnels to validate the planning framework as well as the robustness of the whole system.
\end{abstract}

\section{Introduction}
\label{sec:introduction}

Because of their agility and maneuverability, multirotors, as one of the most ubiquitous types of micro aerial vehicle (MAV), are playing important roles in an abundance of missions, including inspection\cite{mathe2016vision}, search \& rescue\cite{luqi2018collaborative}, and surveillance\cite{manyam2017surveillance}. During these missions, multirotors are desired to enter confined and narrow spaces that are barely accessible to humans. 

\begin{figure}[t]
\begin{center}
\subfigure[\label{fig:tunnel_0} The straight tunnel.]
{\includegraphics[width=0.48\columnwidth]{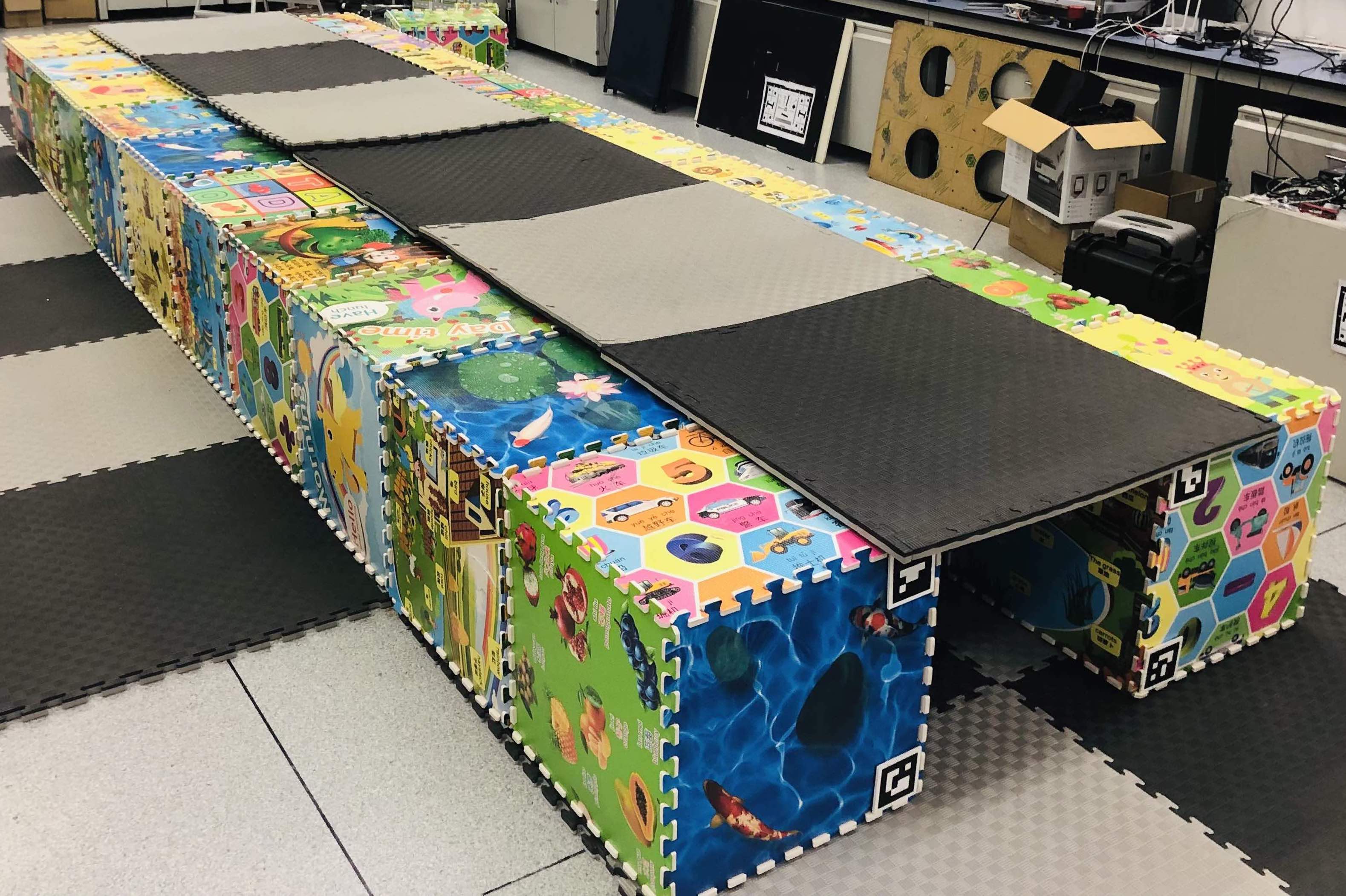}} 
\subfigure[\label{fig:tunnel_3} Curved tunnel case 1.]
{\includegraphics[width=0.48\columnwidth]{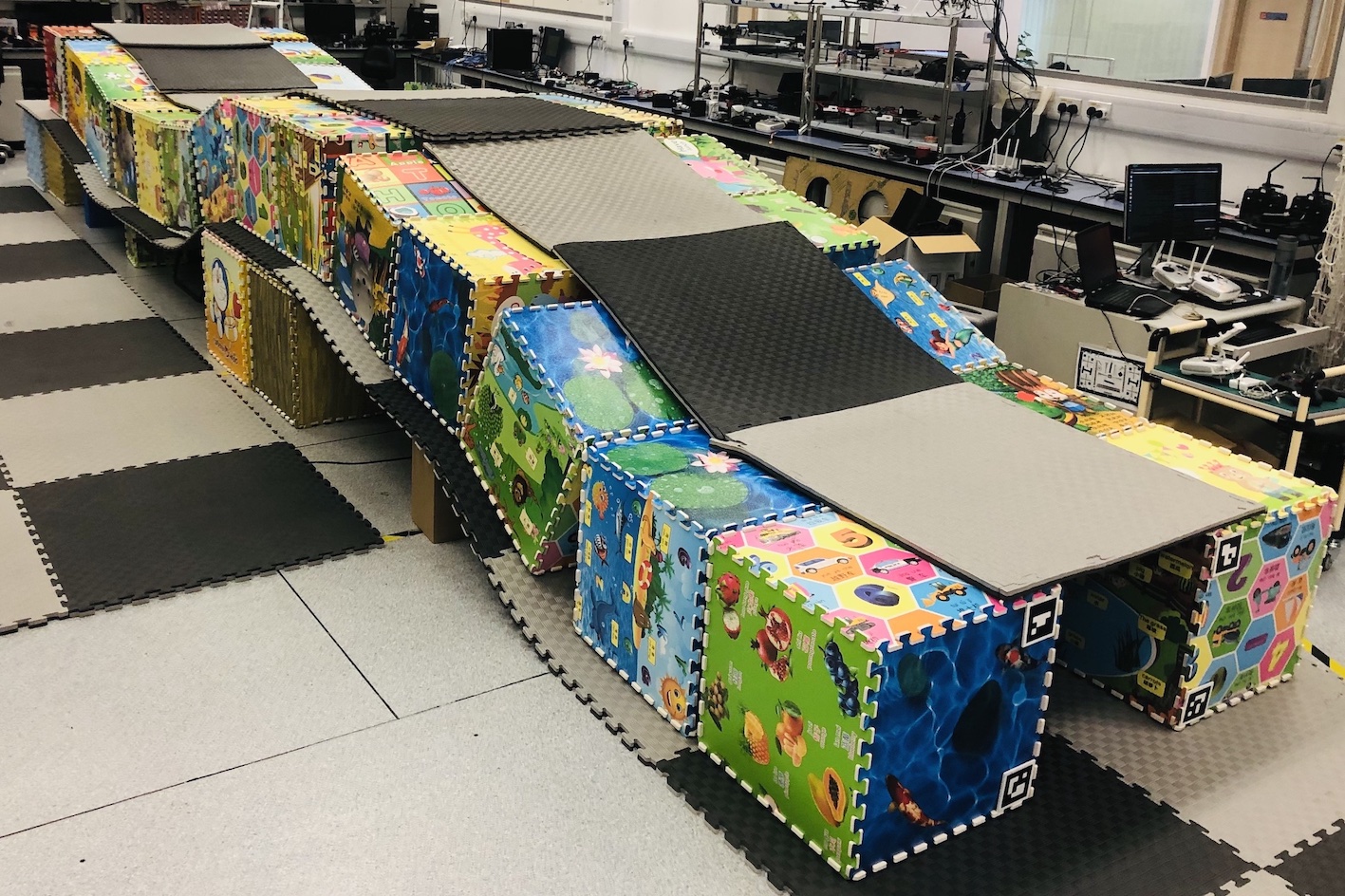}} 
\subfigure[\label{fig:tunnel_1} Curved tunnel case 2.]
{\includegraphics[height=0.32\columnwidth]{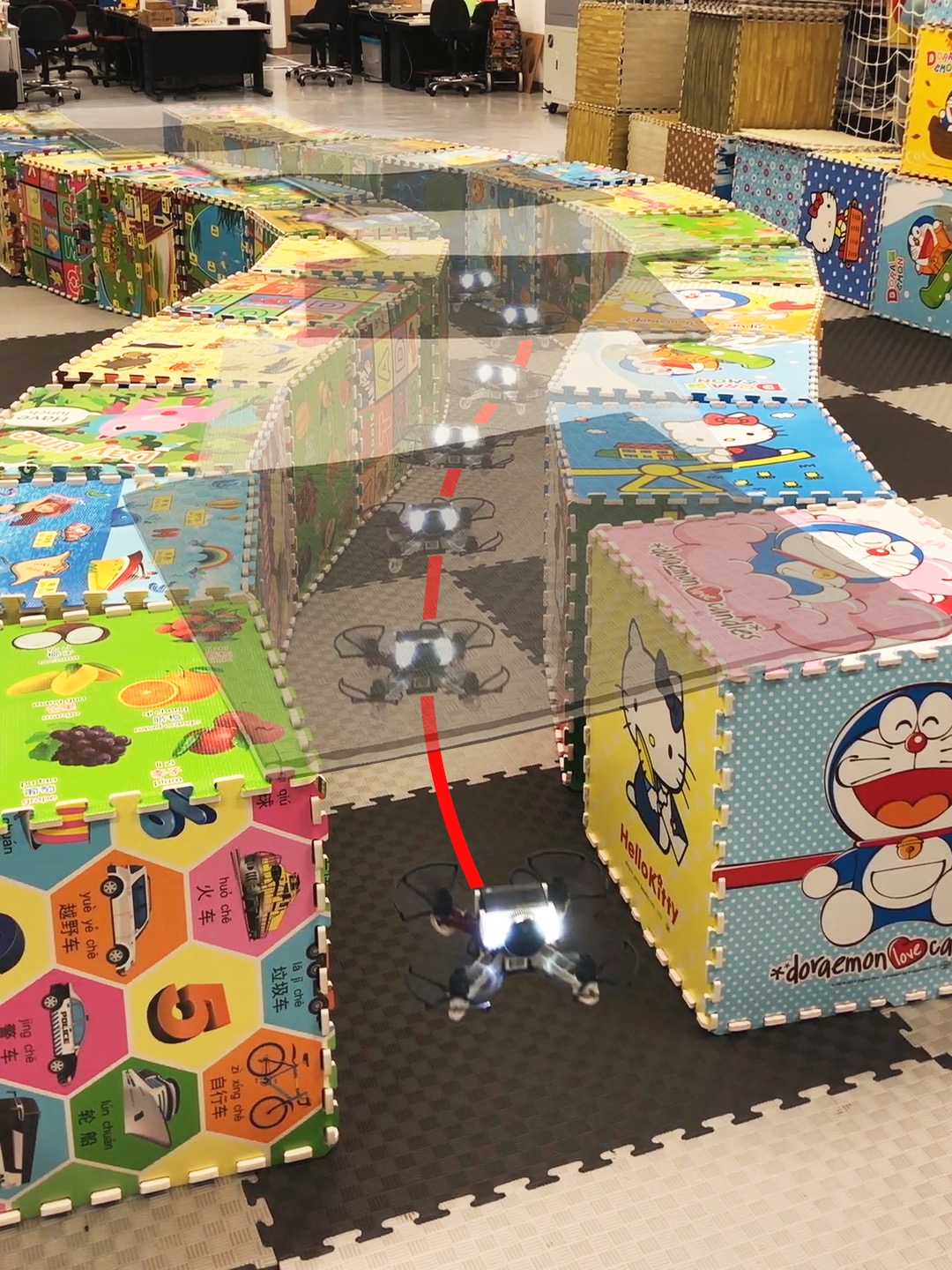}} 
\subfigure[\label{fig:tunnel_2} Curved tunnel case 3.]
{\includegraphics[height=0.32\columnwidth]{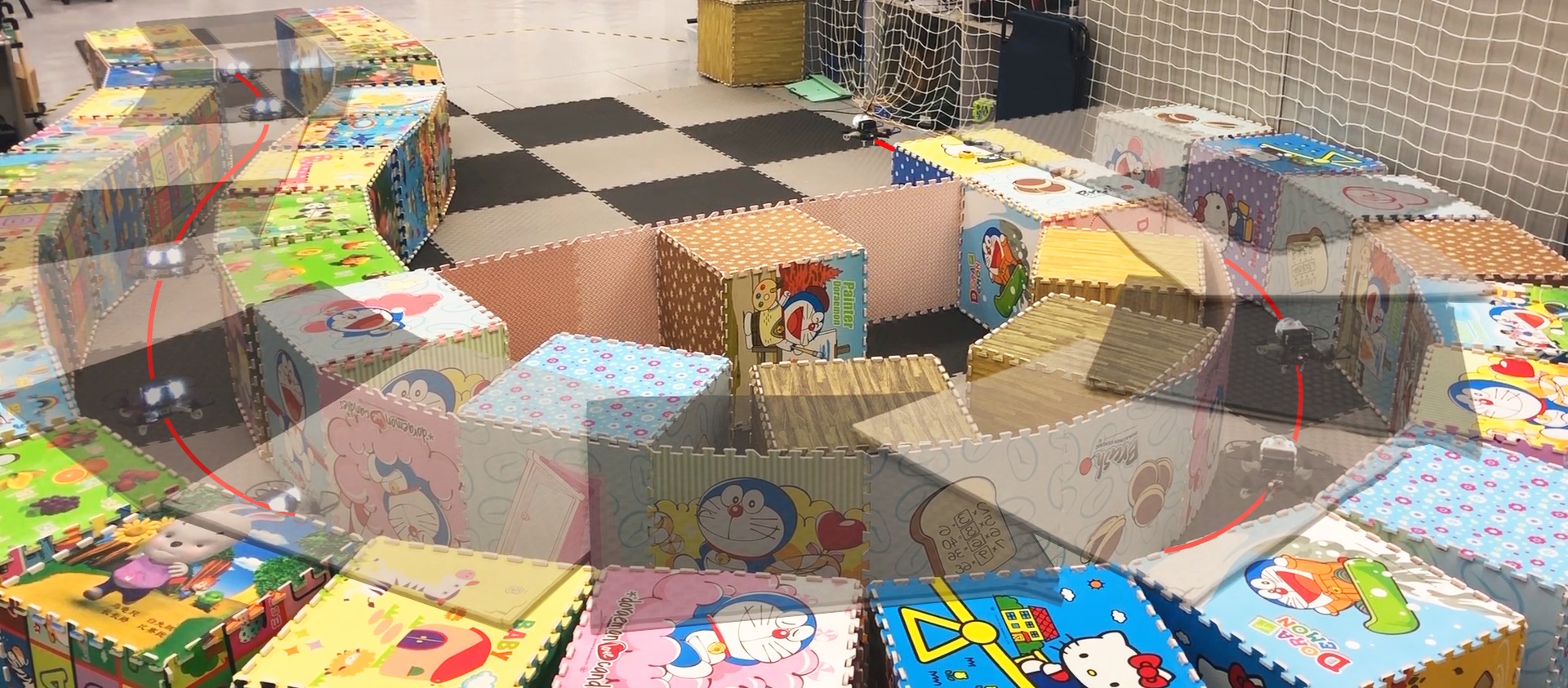}} 
\subfigure[\label{fig:vent_case_1} Vent pipe case 1.]
{\includegraphics[height=0.41\columnwidth]{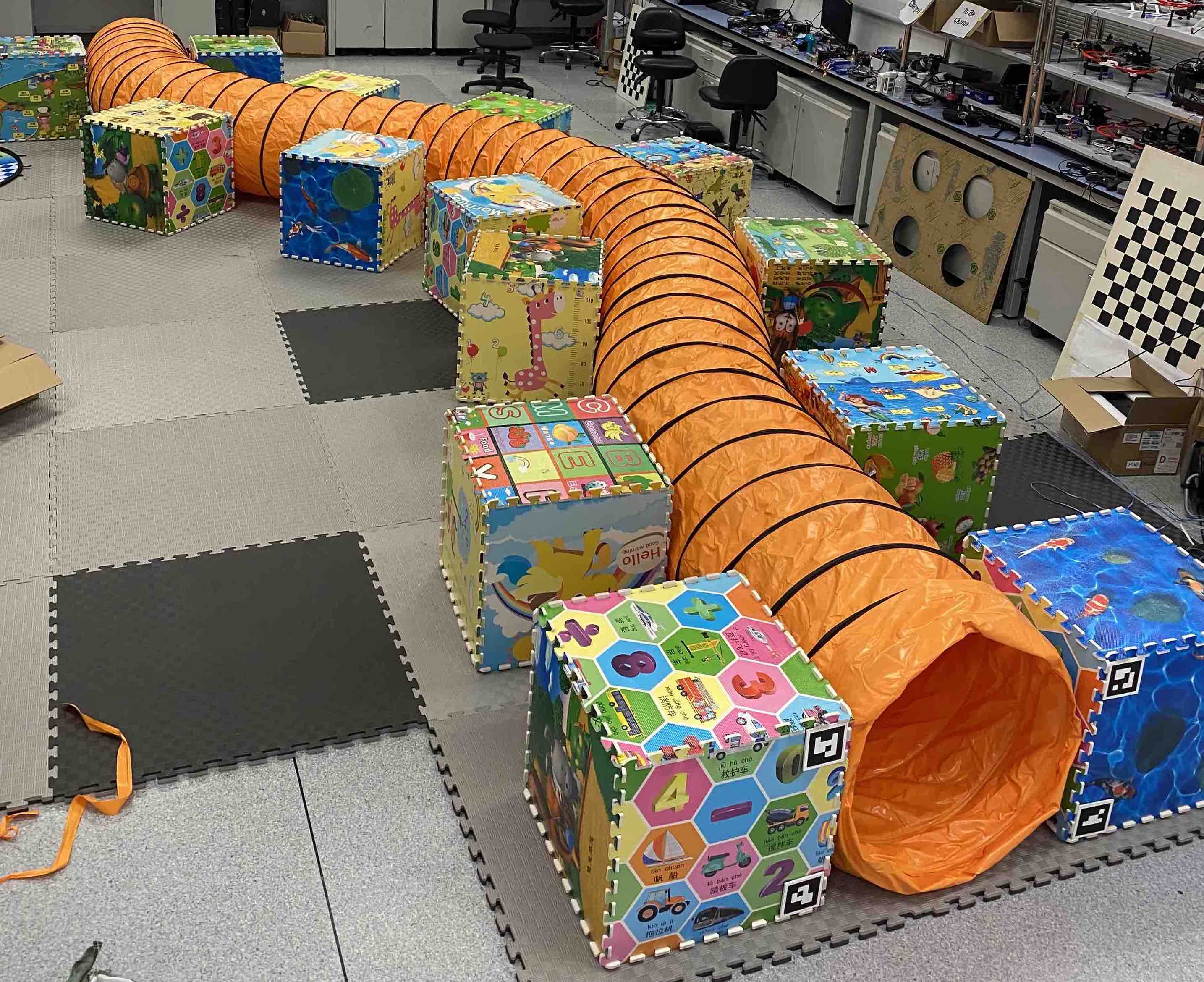}} 
\subfigure[\label{fig:vent} Vent pipe case 2.]
{\includegraphics[height=0.41\columnwidth]{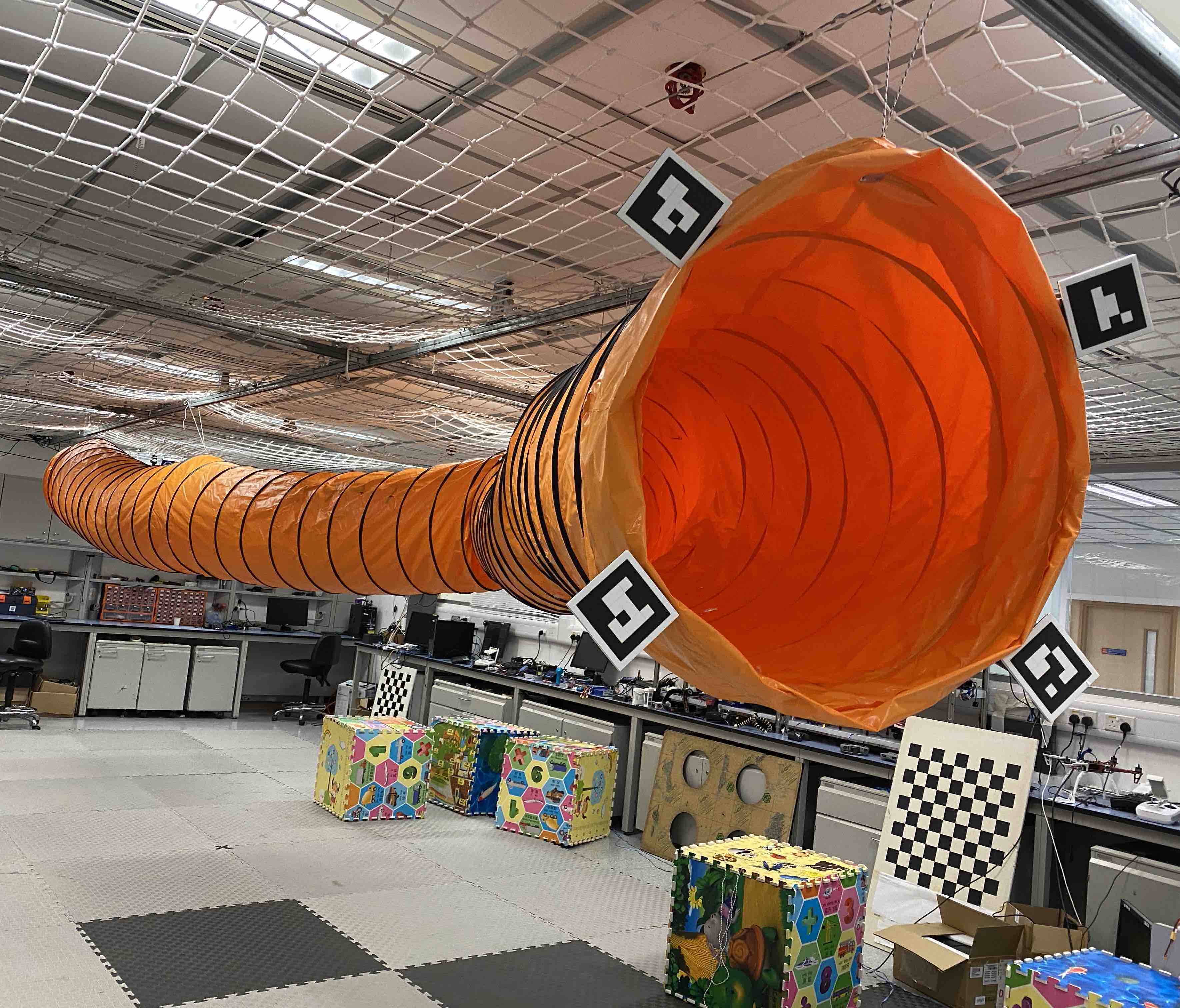}} 
\end{center}
\vspace{-0.4cm}
\caption{\label{fig:tunnel} The narrow tunnels and vent pipes for the flight tests and the composite images indicating the trajectories.}
\vspace{-0.5cm}
\end{figure}

One of the typical yet extremely challenging and barely addressed scenarios that is focussed by this letter is traversing a narrow tunnel-like area, for instance, drainage and ventilation conduits and various other types of pipelines. However, these narrow tunnels are exceptionally tricky for multirotors. As pointed out by the previous works\cite{ozaslan2017autonomous,vong2019integral,wang2021estimation}, the challenges are the following:
 
\begin{itemize}

\item The difficulty in control. The restricted manoeuvring space and the problematic ego airflow disturbances from the proximity effects can be detrimental during flight, even when the multirotor is equipped with a controller possessing decent performance in broader areas.

\item The difficulty in state estimation and perception. Besides the absence of a global positioning system, the lack of geometric features and external illuminations induces unobservability in ranging-based state estimations and the failure of visual state estimations.

\end{itemize}

To compensate for control error caused by large ego airflow disturbances and thus ensure safety, in previous work \cite{wang2021estimation}, the multirotor and the induced control error bounds are kept away from the obstacles. However, in narrow tunnels, the restricted space prevents the multirotor from maintaining a sufficient safety margin, ruling out this solution. In this case, inspired by\cite{cheeseman1955effect}, an intuitive solution is to increase the flight speed to mitigate the disturbances from the proximity effects. Although flying at a high speed can nullify most disturbances caused by the turbulent downwash, the unbalanced forces caused by the proximity effects still exist\cite{jimenez2019contact,conyers2019empirical}. Hence, a motion planner is required for generating a smooth trajectory along the tunnel center line. Nonetheless, even with the commands on the center line, larger flying speeds can also produce larger control errors \cite{morrell2018comparison}, owing to the limited control bandwidth of the propulsion system, easily causing crashes in such narrow spaces. Therefore, a proper flight speed range is essential as well.


Initially, this work purely aimed at developing a planner and figuring out the practical speeds to address the control issues and tried to bypass the state estimation and perception issues. However, since the multirotor needs to fly inside narrow tunnels where external positioning systems are not available, a complete tunnel autonomous flight system including a state estimation and a perception module needs to be developed. In this work, we choose from the state-of-the-art (SOTA) state estimation and perception methods to build up the system. As pointed out in\cite{ozaslan2017autonomous}, LiDAR-based odometries are not applicable since the symmetric geometry of tunnels causes unobservability in the longitudinal direction, while the alternative, visual-inertial odometry (VIO), can be made functional by adding illuminations on the multirotor. To make the state estimation and perception module function properly, a smooth motion planner along the tunnel center line at a practical speed is even more necessary. Despite the illumination by auxiliary lights, the lighting conditions inside tunnels are still not comparable to well-illuminated areas outside. Therefore, to ensure good feature tracking performance of the VIO, which also plays an important role in the perception module, smooth motion commands are required. Even with smooth commands along the tunnel center line, which benefits the VIO and the mapping, flight speeds can still play an important role. On the one hand, it is obvious that fast flights are not favorable to the VIO system, since the large motion blur and parallax occurring at high speeds in such constrained scenarios can engender more unstable feature tracking under the limited illumination\cite{liu2021mba}. On the other hand, at a slow flight speed, the shaky motion induced by airflow disturbances can also result in poor feature tracking performance, affecting both the VIO and the map, and thus further causing control issues. As a result, neither fast nor slow speeds in narrow tunnels are practical, and the compensation is tremendously crucial for safe flights.

In this letter, a complete narrow tunnel autonomous flight system is proposed, and the impacts of different speeds, specifically on the control and state estimation performance are investigated. First, a double-phase motion planner containing center line extraction and trajectory optimization is designed for flights through narrow tunnels. Then, the proposed motion planner is deployed on a customized quadrotor platform, and numerous flight tests are performed in a straight narrow tunnel at various speeds are performed to collect real-time data and further analyzed with the data collected in broader areas to determine the optimal speed range. During the speed selection process, computational fluid dynamics (CFD) analyses are also conducted to validate the intuition about the relationship between speed and disturbance. Moreover, multiple flight experiments in several curved narrow tunnels and vent pipes are conducted to validate the proposed motion planner in more complex situations as well as prove the robustness of the whole system.

The contributions of this letter are the following:
\begin{enumerate}
	\item A double-phase motion planning framework to smoothly fly a multirotor through narrow tunnels.
	\item The optimal speed range for the quadrotor to fly through the narrow tunnels as determined through straight tunnel flight data.
	\item A set of CFD analyses on the validation of the relationship between speed and disturbance.
	\item Comprehensive integration of a narrow tunnel autonomous flight system, including the proposed motion planning method, together with visual-inertial state estimation, mapping, and control modules, into a customized quadrotor platform equipped with illuminations. Extensive narrow tunnel flights are performed to verify the planning method and the robustness of the entire system, meanwhile collecting data for analyses. \textbf{To the authors' best knowledge, this is the first autonomous multirotor system that can fly through tunnels narrow to 0.6m.}
\end{enumerate}

\section{Related Work}
\label{sec:related_work}
\textbf{Multirotor flights in tunnel-like confined areas:}
Multirotor flights in constrained and cluttered environments have been studied for years, and a large amount of motion planning frameworks, together with system integration, have been proposed \cite{chen2016online,loianno2016estimation,zhou2019robust,gao2020teach}. However, flights in tunnel-like confined areas are not as comprehensively studied. As mentioned in \cite{ozaslan2017autonomous}, the state estimation inside tunnels can be challenging due to the lack of light and geometry features. Additionally, multiple proximity effects, including the ground effect, which repels the multirotor from the ground, and the near-wall effect and ceiling effect, which attract the multirotor towards the ceiling and walls \cite{powers2013influence,eberhart2017modeling,conyers2019empirical} combine to form extremely complex scenarios that bring about severe control issues\cite{vong2019integral}. In \cite{elmokadem2021method}, another navigation method with system integration is proposed. However, since the planned trajectories are not smooth enough, even in tunnels with a large diameter of several meters, the control performance may still be unsatisfactory. Therefore, it is likely that in narrow tunnels with dimensions narrow to 0.6 m, which are the focus of this work, the performance would be even worse and the situation exceptionally challenging.

\textbf{Speed effects during multirotor flights:}
In general, flying at a high speed is considered by previous works to be more difficult. During high-speed flights, the feature tracking in the state estimation system becomes less stable and the requirement for low-latency computation time is not easy to achieve\cite{shen2013vision}. Additionally, due to the unstable tracking and the physical limitation of the multirotor, the control error for tracking a high-speed trajectory becomes larger \cite{morrell2018comparison,fridovich2018planning}, which can be dangerous when flying near obstacles, and may even result in a collision. As a result, multiple motion planning approaches that adapt speeds according to the distance from obstacles, i.e. flying faster in broader areas and slower in narrow areas, have been proposed \cite{fridovich2018planning,quan2020eva}. However, the ego airflow disturbances become more serious near obstacles, especially during slow flights, leading to a large control error \cite{wang2021estimation}. Without considering these disturbances can also be disastrous.

\section{System and Workflow for Tunnel Flights}
\label{sec:system}


\subsection{System Architecture}
\label{subsec:sys_arch}

\begin{figure}[t]
\begin{center}
\subfigure[\label{fig:drone}The customized quadrotor platform.]
{\includegraphics[width=0.8\columnwidth]{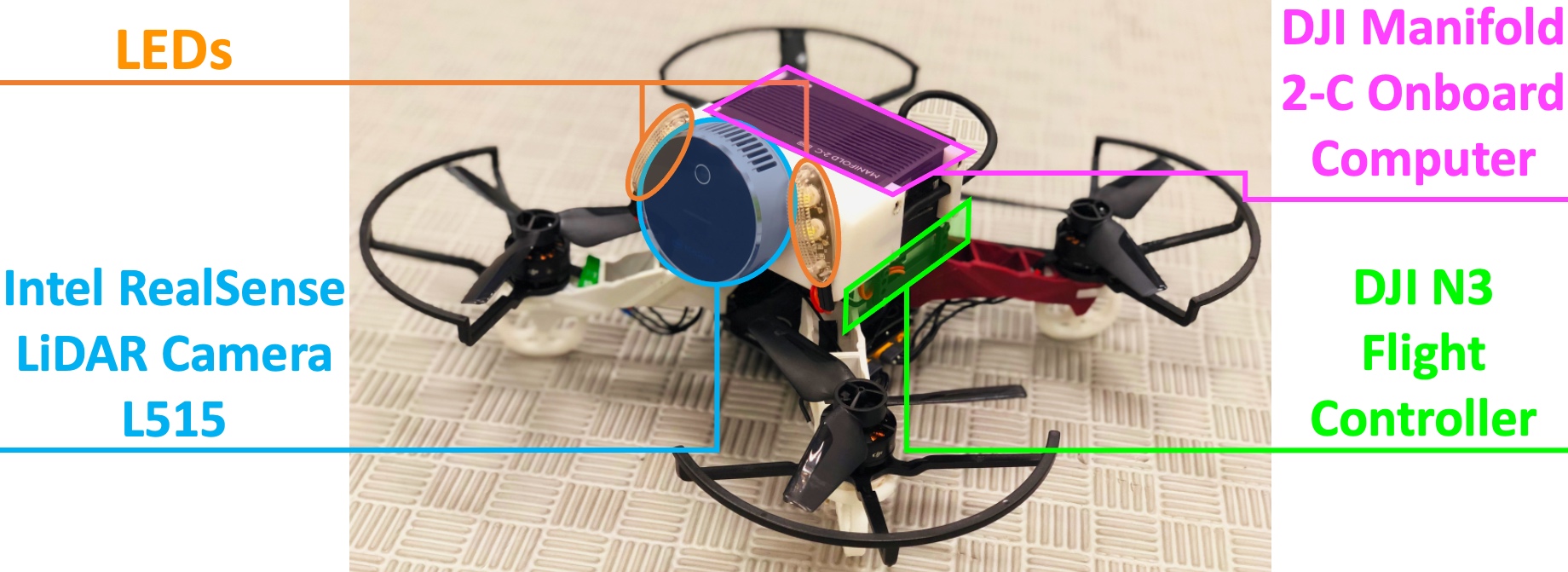}}             
\subfigure[\label{fig:system}The system architecture and the workflow.]
{\includegraphics[width=0.85\columnwidth]{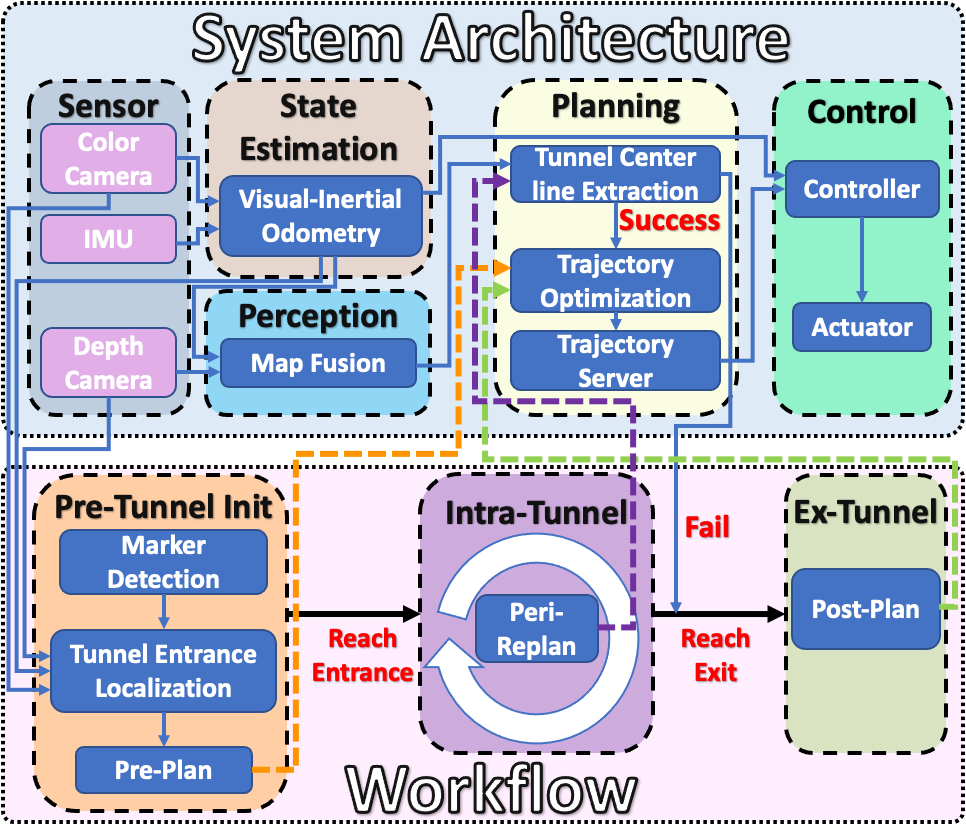}} 
\end{center}
\vspace{-0.3cm}
\caption{\label{fig:hard_soft}The hardware and the software system architecture, together with the workflow for the quadrotor to fly through narrow tunnels.}
\vspace{-1.2cm}
\end{figure}

To facilitate flight inside narrow tunnels, a customized quadrotor platform, as shown in Fig. \ref{fig:drone}, is designed. Owing to the absence of external localization systems, an onboard state estimation and mapping system is required for safe flight. Due to the lack of geometry features and the complete lack of external illumination, as mentioned in Sec. \ref{sec:introduction}, VIO together with additional LEDs is adopted for the state estimation. For the perception module, a relatively light-insensitive LiDAR is a practical solution on account of the abrupt change in lighting conditions at the entrance and the exit of a tunnel, bringing about severe issues for vision-based depth estimation, further corrupting the map fusion. Our system uses an Intel RealSense LiDAR camera L515 consisting of both a LiDAR and a color camera as the main perception sensor. As indicated in Fig. \ref{fig:hard_soft}, a software system consisting of state estimation, perception, planning, and high-level control, is integrated into a DJI Manifold 2-C onboard computer. The SOTA VIO, VINS-Mono \cite{qin2018vins} is adopted for state estimation, while Fiesta \cite{han2019fiesta} is adopted for the map fusion along with Euclidean distance field (EDF) update for the planning module. The control commands are derived by the high-level controller according to the planned trajectory as well as the VIO, and are further sent to the DJI N3 flight controller for execution by the motors.

\subsection{Tunnel Flight Workflow}
\label{subsec:workflow}
\begin{figure}[t]
\begin{center}
\includegraphics[width=0.6\columnwidth]{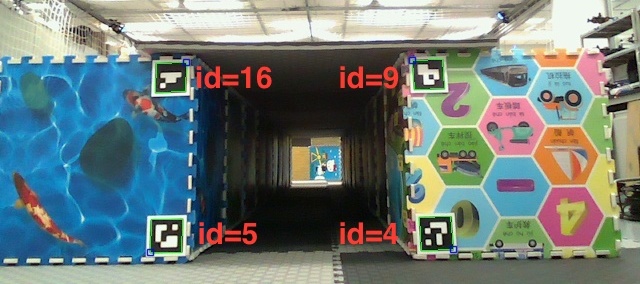}
\end{center}
\vspace{-0.4cm}
\caption{\label{fig:marker}The detected ArUco markers at the tunnel entrance and the corresponding marker IDs.}
\vspace{-0.9cm}
\end{figure}

To ensure safe and smooth tunnel flights, we design a workflow as shown in Fig. \ref{fig:hard_soft}. Since the sudden change in aerodynamics at the tunnel entrance can cause unsteady motion in flight, a more precise pose of the center of the tunnel entrance is desirable to minimize the disturbance. As shown in Fig. \ref{fig:marker}, four ArUco markers can be easily installed symmetrically at the tunnel entrance for entrance localization during the pre-tunnel initialization. By combining the detected 2-D marker positions on a color image and the corresponding depths from the depth camera, adding to the camera pose estimated from the VIO, the 3-D marker positions can be calculated. Multiple detections are performed and the outliers are rejected to obtain more accurate 3-D positions. The position of the entrance is obtained as the mean of the four marker positions, while the direction of the tunnel entrance is derived by solving the normal of the least-squares error plane determined by the four marker positions using singular value decomposition (SVD). A mini-jerk trajectory \cite{mellinger2011minimum} is generated towards the entrance with a target ending velocity, where the magnitude is the desired speed inside the tunnel and the direction is aligned with the tunnel entrance. When the quadrotor reaches the entrance, the state will switch to intra-tunnel and the peri-replanning that keeps the quadrotor on the center line according to the updating map is performed at 10 Hz until the quadrotor reaches the exit. Details about the planning will be introduced in Sec. \ref{subsec:planner}. At the moment the exit is reached, a post-planning for deceleration is carried out also utilizing the mini-jerk trajectory.

\subsection{Double-phase Motion Planning in Narrow Tunnels}
\label{subsec:planner}

\subsubsection{Tunnel Center Line Extraction}
\label{subsubsec:center_line_extraction}

\begin{algorithm}[t]
\caption{Tunnel Center Line Extraction}  
\label{alg:center_line}  
\begin{algorithmic}
\State \textbf{Notation}: Start point $P$, Start velocity $V$, Waypoints $\mathcal{W}$, Point $p$, Tunnel dimension $D$, Search step length $S$, EDF value $d$, Point set $\mathcal{P}$, Plan distance $d_p$, Plan range $R_p$, Tunnel center line trajectory $T$
\State \textbf{Input}: $P$, $V$
\State \textbf{Output}: $T$
\State Initialize :
\State $d_p \leftarrow 0$
\State $dir \leftarrow V$.normalize()
\State $\mathcal{W}$.push\_back($P$)
\State $p \leftarrow GradientAscend( P + S \cdot dir, dir)$
\While{$p.d \leq 0.5 \cdot D \ \&\& \  d_p \leq R_p$}
\State $\mathcal{W}$.push\_back($p$)
\State $\mathcal{P} \leftarrow SphereRandomSample(p)$
\ForEach {$p_i \in \mathcal{P}$}
\State $p_i \leftarrow SphereGradientDescend(p_i)$
\EndFor
\State $dir \leftarrow PlaneFit(\mathcal{P}, dir).normal()$
\State $p \leftarrow GradientAscend( P + S \cdot dir, dir)$
\State $d_p \leftarrow d_p + dist(p,\mathcal{W}.back())$
\EndWhile
\State $T \leftarrow Bspline(\mathcal{W})$
\State \Return $T$
\end{algorithmic} 
\end{algorithm} 

Since the space inside a narrow tunnel is constrained, the proximity effects are significant. As mentioned in \cite{eberhart2017modeling,conyers2019empirical,wang2021estimation}, the proximity effects, i.e. the ground effect, the near-wall effect, and the ceiling effect, can generate not only additional mean forces but also disturbances. In narrow tunnels, turbulent downwashes bounce back and forth, inducing complicated and unpredictable effects rather than a simple superposition of the proximity effects, and thus bring tremendous difficulty in flight. With consideration to these effects, it is crucial to reserve as large a clearance as possible from the tunnel walls; thus, following the center line is a desirable resolution. Additionally, on account of the 0.25m minimum working range of the LiDAR, keeping enough clearances from the walls is also beneficial to the perception module, further justifying the solution.

The algorithm for the tunnel center line extraction is shown in Alg. \ref{alg:center_line}. During this process, the local EDF keeps updating according to the fused map to facilitate the following procedures. With the assumption of constant tunnel dimensions $D$ and the search step length $S$, the waypoints on the tunnel center line can be extracted from the start point $P$ and the start velocity $V$, which are determined from the currently executing trajectory command. For each iteration, firstly, the gradient ascent of the point $p$ is performed in the plane normal to the current direction $dir$ according to the EDF value. Then, when $p$ reaches the position with the local maximum EDF value, eight random samples on the sphere centered at $p$ with the radius of the EDF value at $p$ are generated in each of the corresponding octants. After that, gradient descent of the sampled points $p_i$ pertaining to the EDF value is performed to move the points towards the positions nearest to the tunnel surfaces. Finally, the forward step direction is obtained through the least-squares plane fitting of the eight points using SVD. The loop is repeated until the EDF value at $p$ is greater than the radius of the tunnel or the planned distance reaches the maximum range. When the loop reaches the end, the waypoints are parameterized into a B-spline trajectory according to the desired speed.

\subsubsection{Trajectory Optimization}
\label{subsubsec:traj_opt}
The extracted B-spline tunnel center line is generally jerky due to the perception noise, and thus cannot be directly used for a smooth flight. Therefore, we need to generate an optimized smooth trajectory at the desired speed from the center line for the quadrotor to execute. A trajectory optimization method extended from our previous work\cite{zhou2019robust} is proposed.

The total cost function $f_{total}$ is defined as the weighted sum of the smoothness cost $f_s$, the waypoint constraint cost $f_w$, the speed constraint cost $f_v$, and the initial and end state constraint costs $f_i$ and $f_e$:
\begin{equation}\label{equ:cost}
	f_{total} = \lambda_{s} f_{s} + \lambda_{w} f_{w} + \lambda_{v} f_{v} +  \lambda_{i} f_{i} +  \lambda_{e} f_{e}.
\end{equation}

The smoothness cost $f_s$ is set to be the elastic band cost \cite{quinlan1993elastic,zhu2015convex} approximating the third-order derivatives of the control points, and is closely related to the jerk cost on the trajectory:
\begin{equation}\label{eq:elastic} 
	f_{s} 
	= \sum\limits_{i=0}^{N-p_b} \Vert -\mathbf{Q}_{i} + 3\mathbf{Q}_{i+1} - 3\mathbf{Q}_{i+2} + \mathbf{Q}_{i+3} \Vert^{2},
\end{equation}
where $p_b$ is the order of the B-spline and $p_b \geq 3$, $\mathbf{Q}_{i}$ represents the position of the $i$-th control point, and $N$ represents the total number of control points.

The waypoint constraint cost $f_w$ is defined as the summation of the squared deviations of the waypoints $W_i$ to ensure the optimized trajectory follows the center line:
\begin{equation}\label{eq:waypoint} 
	f_{w} 
	= \sum\limits_{i=0}^{N-p_b} \Vert EvalBspline(\mathbf{Q}_{i},...,\mathbf{Q}_{i+p_b-1}) - W_{i} \Vert^{2},
\end{equation}
where the $EvalBspline$ function evaluates the waypoint on the B-spline according to the corresponding control points.

The speed constraint cost $f_v$ is penalized on the control points deviating from the desired speed $v_{des}$ in order to maintain constant desired speed during flight:
\begin{equation}\label{eq:speed}
	f_{v} = \sum\limits_{i=0}^{N-1} (\Vert \frac{\mathbf{Q}_{i+1} - \mathbf{Q}_{i}}{\delta t} \Vert  - v_{des})^2,
\end{equation}
where $\delta t$ is the time interval between the control points.

The initial and end state constraint costs $f_i$ and $f_e$ are added according to the difference between the states on the trajectory to be optimized and the original initial and end states:
\begin{equation}\label{eq:start}
	f_{i} = \sum\limits_{i=0}^{k} \Vert EvalBspline_i(\mathbf{Q}_{0},...,\mathbf{Q}_{p_b-1}) - I_i \Vert^{2},
\end{equation}
\begin{equation}\label{eq:end}
	f_{e} = \sum\limits_{i=0}^{k} \Vert EvalBspline_i(\mathbf{Q}_{N-p_b},...,\mathbf{Q}_{N-1}) - E_i \Vert^{2},
\end{equation}
where the $EvalBspline_i$ function evaluates the $i$-th derivative of the point on the B-spline according to the corresponding control points, and $I_i$ and $E_i$ indicate the $i$-th derivative of the original initial state and ending state, respectively.
In practice, the weight of the initial state cost $\lambda_{i}$ is chosen to be larger than the other weights to achieve smoothness at the start of the trajectory.

Even with soft constraints of the initial state, the initial state on the optimized B-spline still differs slightly from the original initial state. Hence, a hard constraint is enforced using the mini-jerk trajectory generator\cite{mellinger2011minimum}. Subsequent to the initial state, the waypoints for generating the final trajectory are selected on the optimized B-spline with a constant time interval. The final smooth mini-jerk trajectory along the tunnel center line is sent to the trajectory server and then to the controller for execution.

\subsection{Speed Range Selection}
\label{subsec:speed_select}

\subsubsection{CFD Analysis}
\label{subsubsec:cfd}
\begin{figure}[t]
\begin{center}
{\includegraphics[width=0.5\columnwidth]{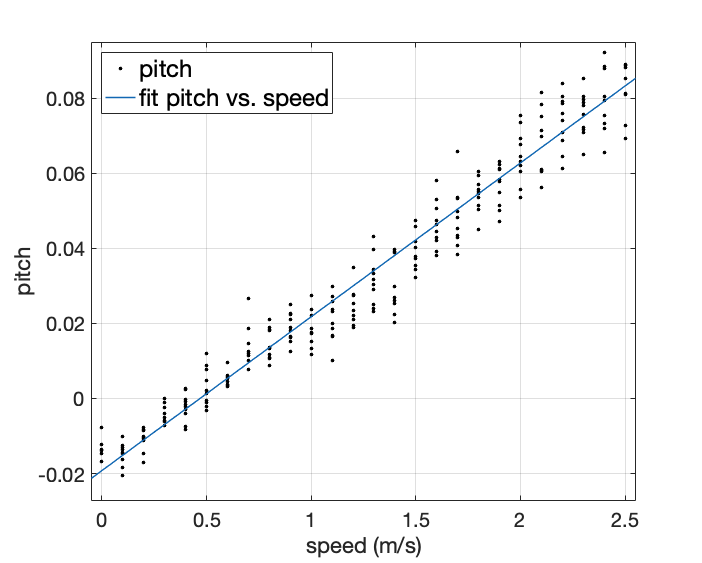}} 
\end{center}
\vspace{-0.4cm}
\caption{\label{fig:pitch}The pitch data and the fit line against the flight speed in broad areas. The slope of the line is 0.041.}
\vspace{-1.5cm}
\end{figure}

It is intuitive that flying at higher speeds can mitigate the effect of ego airflow disturbances. To validate this intuition, a set of CFD analyses at 6 different speeds of 0.2 m/s, 0.5 m/s, 1 m/s, 1.5 m/s, 2 m/s and 2.5 m/s is conducted according to the experiment settings. For a near-hovering 1.23 kg quadrotor with 5-inch propellers, the model can be simplified as four 5-inch fans with a pressure jump of 240 Pa with the thin fan approximation. The pitch angles for constant speed flights are obtained from straight-line flights outside the tunnel, as mentioned in Sec. \ref{subsec:exp_setup}, while the pitch results are shown in Fig. \ref{fig:pitch}. The pitch data are linear to the speed with a slope of 0.041, which coincides with the rotor drag theory \cite{faessler2017differential}. The y-intercept not being zero is possibly because of manufacturing or installation error.

\begin{figure}[h]
\begin{center}         
\subfigure[\label{fig:02_cfd} 0.2 m/s.]
{\includegraphics[width=0.4\columnwidth]{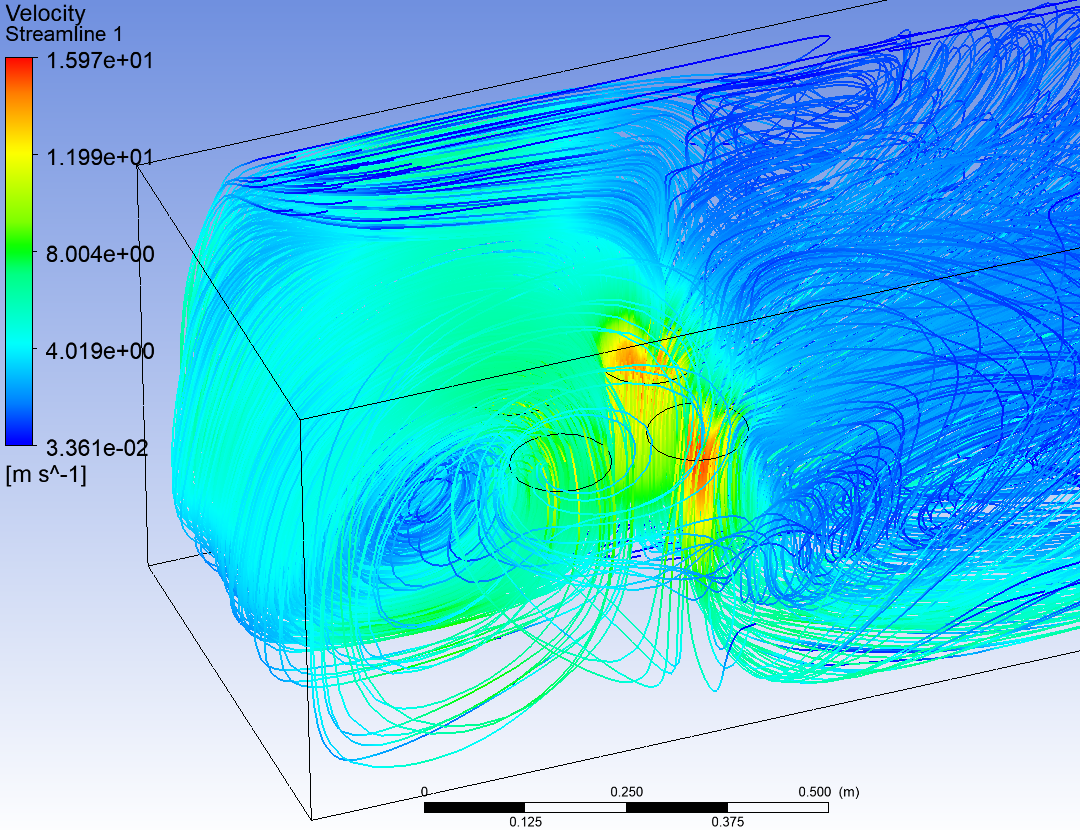}}
\subfigure[\label{fig:05_cfd} 0.5 m/s.]
{\includegraphics[width=0.4\columnwidth]{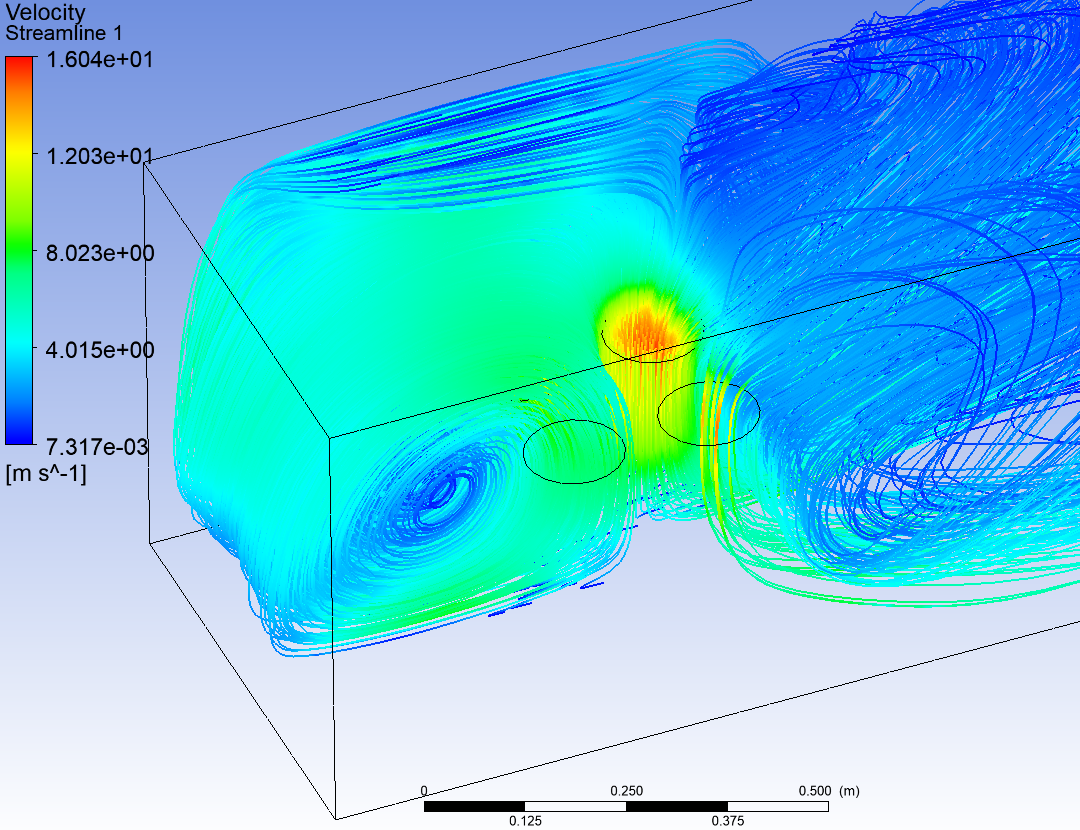}}
\subfigure[\label{fig:10_cfd} 1.0 m/s.]
{\includegraphics[width=0.4\columnwidth]{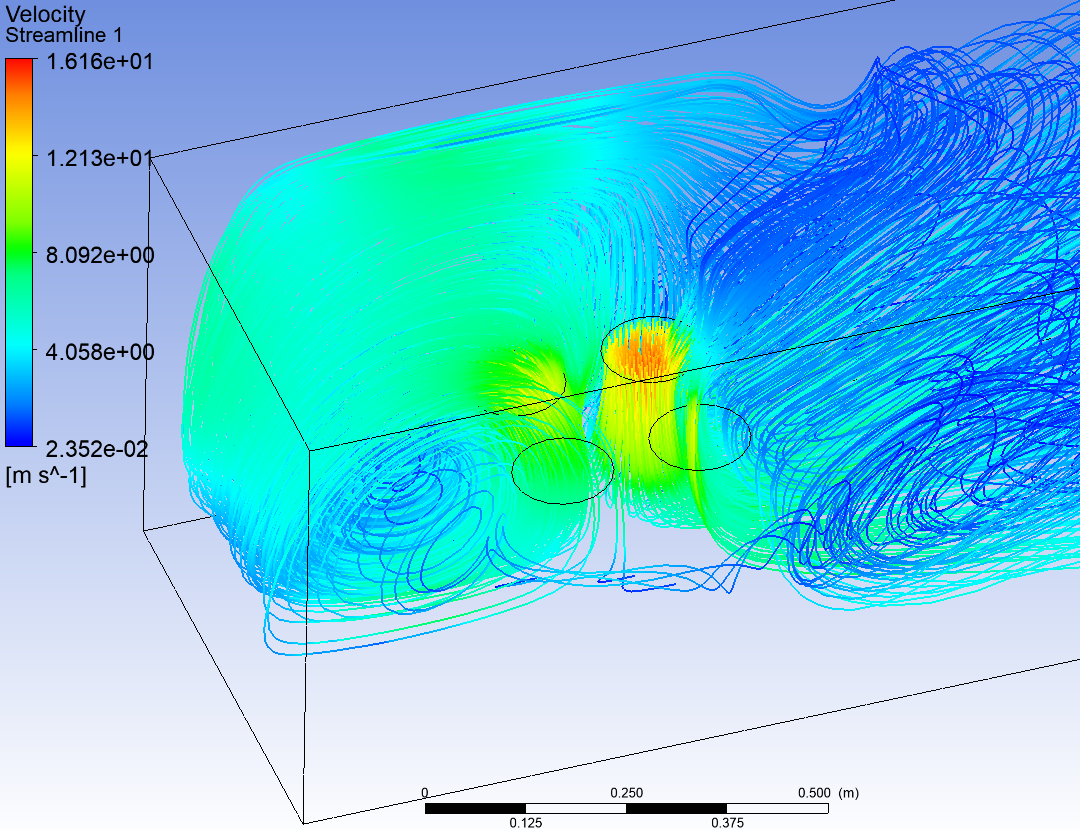}}
\subfigure[\label{fig:15_cfd} 1.5 m/s]
{\includegraphics[width=0.4\columnwidth]{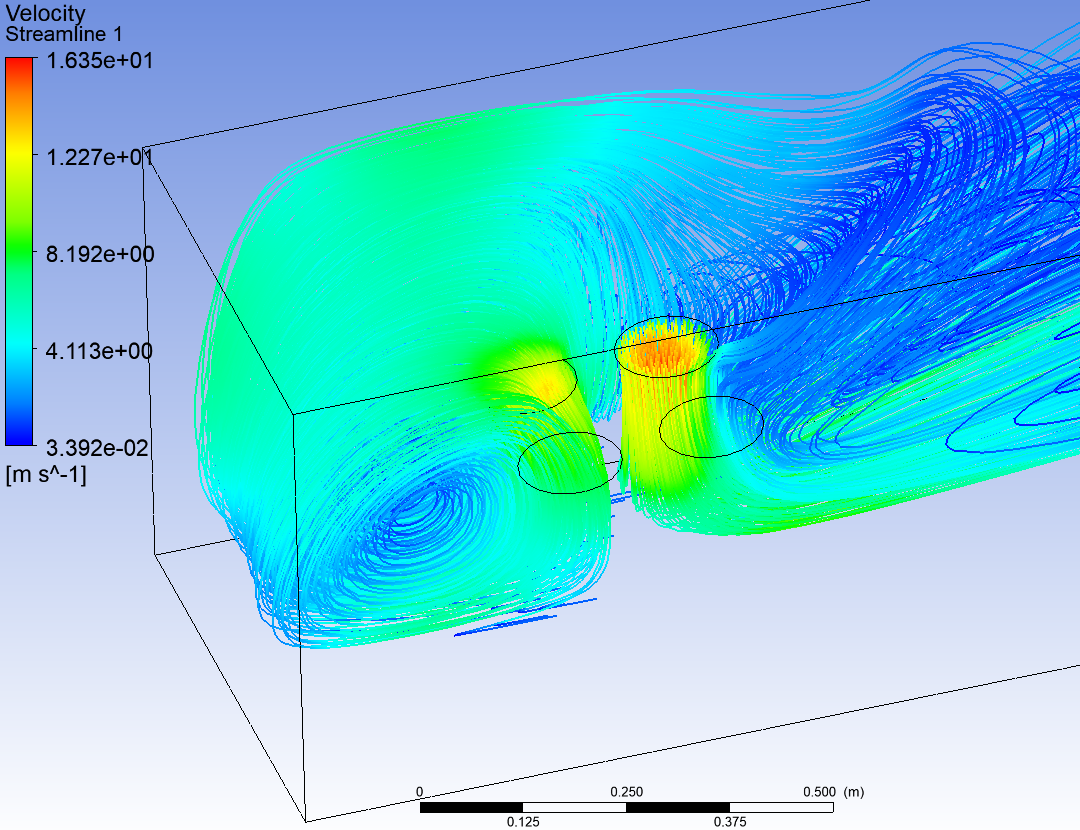}}
\subfigure[\label{fig:20_cfd} 2.0 m/s]
{\includegraphics[width=0.4\columnwidth]{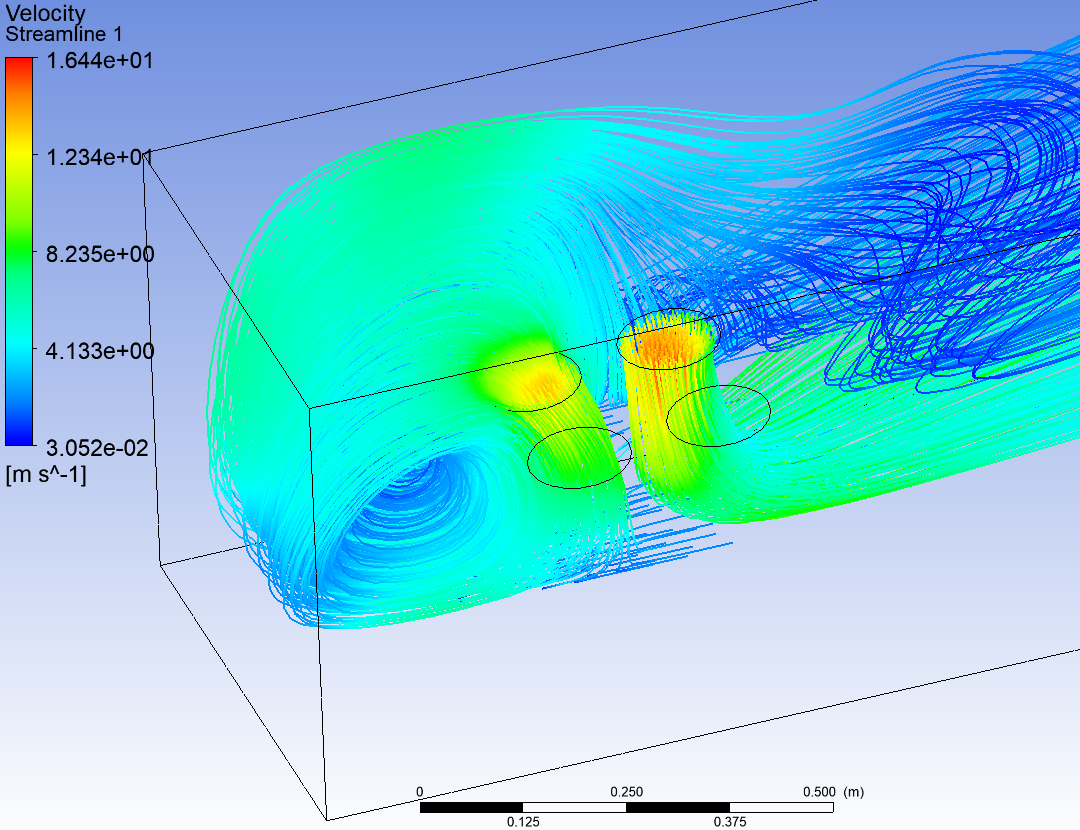}}
\subfigure[\label{fig:25_cfd} 2.5 m/s]
{\includegraphics[width=0.4\columnwidth]{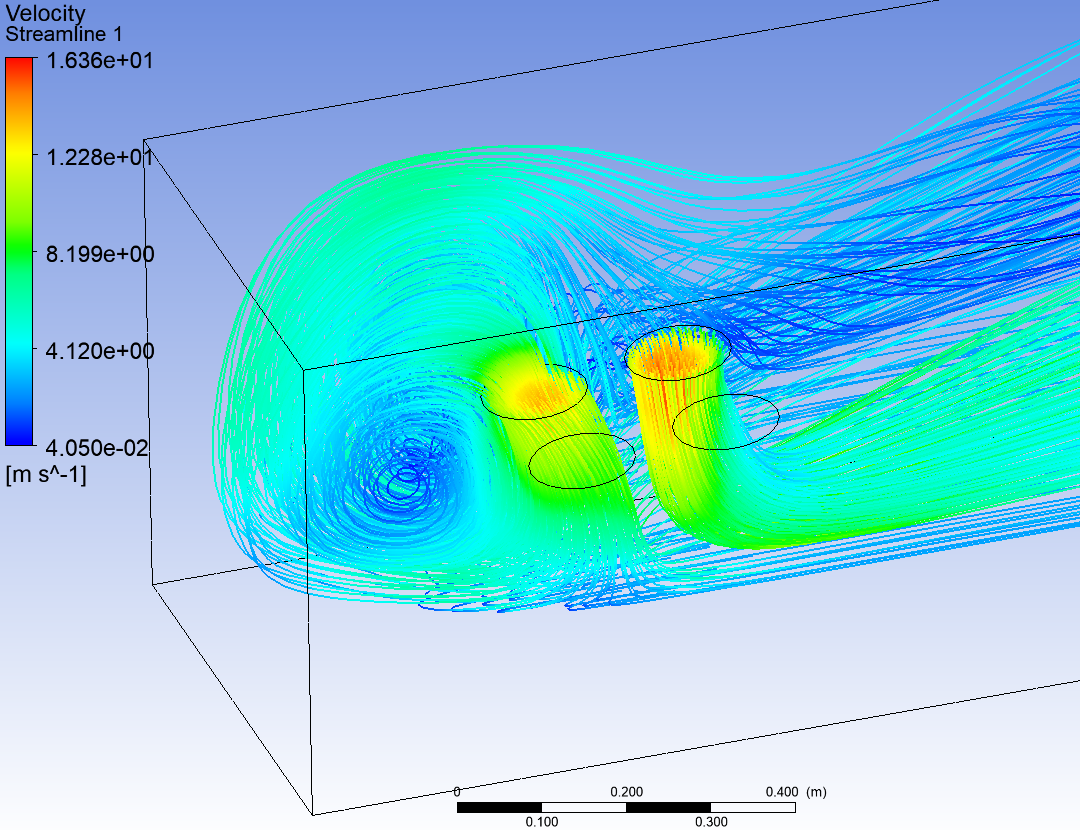}}
\end{center}
\vspace{-0.4cm}
\caption{\label{fig:cfd_result}The forward streamlines from the shadow of the front-right propeller and the back-right propeller from the CFD result. The color indicates the speed of the flow on the streamlines.}
\vspace{-0.5cm}
\end{figure}

The CFD simulations aim to examine the quatrotor flying from right to left in the tunnel, while with the moving frame setup, the airflow moves towards the right from the inlet through the static quadrotor. As shown in Fig. \ref{fig:cfd_result}, four fans representing the quadrotor are placed in a tunnel of 0.6 m in diameter and 2 m  in length, at a distance of 0.5 m from the inlet. The flow velocity at the inlet is set to be the different flight speeds during each simulation, while the outlet is set to be the outlet vent with a restriction of backflows. The walls are set to be the default wall conditions with friction, moving towards the right at the flight speed. The pressure-based, standard k-epsilon turbulence and standard wall model, together with the standard SIMPLE algorithm CFD solver in Ansys Fluent, are adopted to solve the simulation problem.



The CFD results of the flights at six different speeds are shown in Fig. \ref{fig:cfd_result}. In consideration of the symmetry, the figures only depict the forward streamlines from the 250 samples with equal distance on the shadow of the front-right propeller and the back-right propeller each. It can be clearly seen that with the increase of speed, turbulent flows from the propellers are gradually shaken off by the quadrotor, which can reduce the disturbances from the ego airflow. 

\subsubsection{Speed Selection Workflow}
\label{subsubsec:speed_workflow}
Although the CFD analysis can verify the intuition that flying at higher speeds can mitigate the effect of ego airflow disturbances, the errors from modeling, discretization, and so on, as well as complex scenarios in real-world environments, make it hard to generate precise quantitative results for direct usage. Therefore, an experiment-based speed selection workflow is still necessary. Experiments for speed selection are conducted in a straight narrow tunnel of around 0.6 m in diameter, as shown in Fig. \ref{fig:tunnel_0}. The proposed autonomous tunnel flight system traverses the tunnel with desired speeds from 0.1 m/s to 2.5 m/s, with an interval of 0.1 m/s. The flight at each desired speed is repeated 10 times for data collection. The control error data are compared with data of straight-line flights recorded at the same flight speeds in a broader area outside the tunnel. Additionally, the feature tracking data of the VIO system is also collected during the tunnel flights for further analyses. Finally, the practical speed range is selected according to the root-mean-square errors (RMSEs) of the positions and the minimum number of features tracked by the VIO system.


\section{Experiment and Result}
\label{sec:result}

\subsection{Experiment Setup}
\label{subsec:exp_setup}

Experiments are conducted in multiple narrow tunnels of 0.6 m in diameter and two vent pipes of 0.7 m in diameter, as shown in Fig. \ref{fig:tunnel}. The customized 1.23 kg quadrotor platform with 5-inch propellers shown in Fig. \ref{fig:drone} has a diameter of 40 cm, meaning that there is only around 10 cm of clearance on each side in the narrow tunnels and vent pipes.

\begin{figure}[t]
\begin{center}
{\includegraphics[width=0.55\columnwidth]{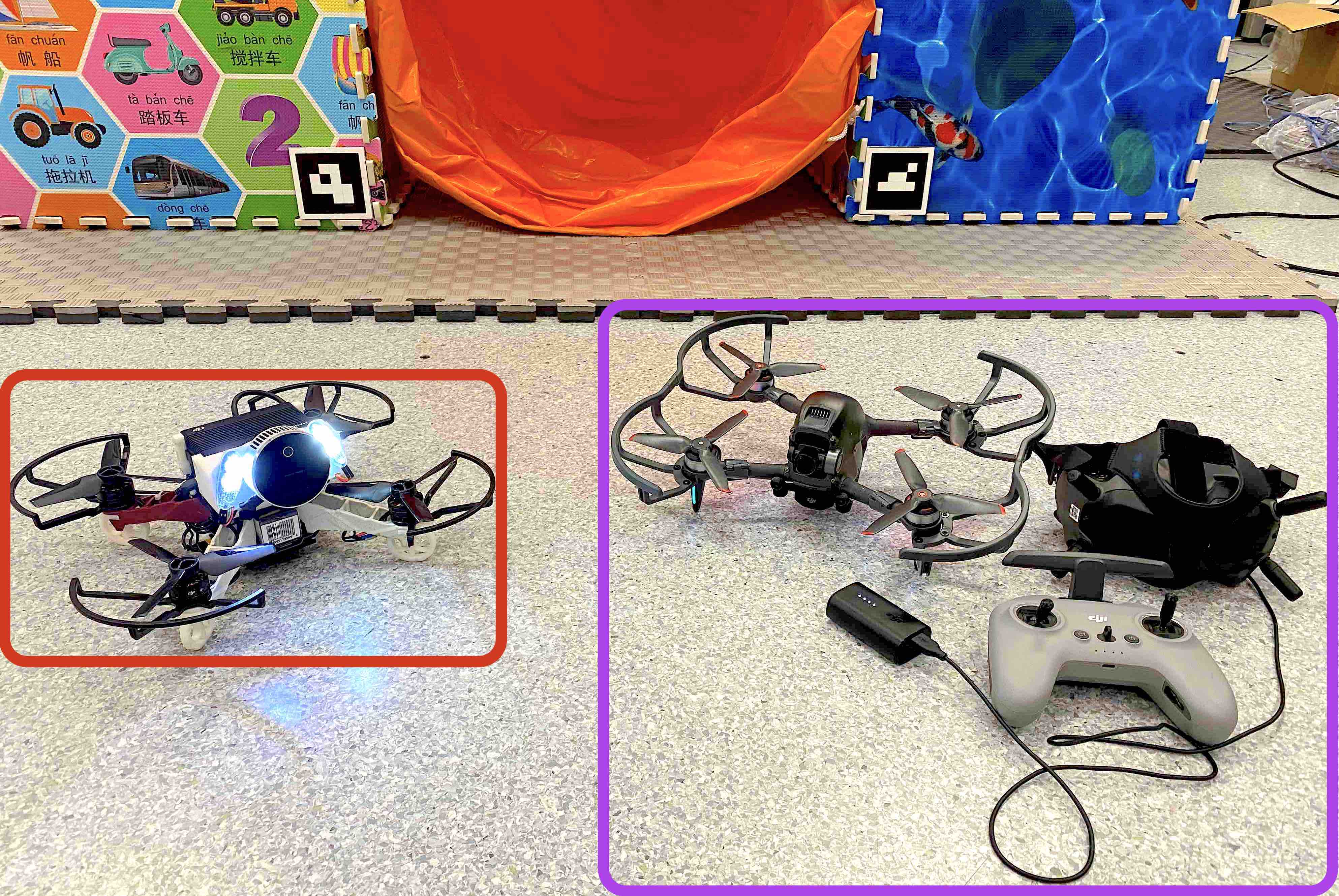}} 
\end{center}
\vspace{-0.3cm}
\caption{\label{fig:drone_fpv}The autonomous tunnel flight system and the commercial FPV system for manual flights, which are shown in the red and the purple frame.}
\vspace{-0.3cm}
\end{figure}

The first set of experiments is conducted in a 6 m-long straight tunnel for speed selection as stated in Sec. \ref{subsubsec:speed_workflow}.
The second set of experiments is conducted in three differently curved tunnels, as shown in Fig. \ref{fig:tunnel_3}, \ref{fig:tunnel_1}, and \ref{fig:tunnel_2}. The flight speeds are set to be 0.2 m/s, 0.5 m/s, 1 m/s, 1.5 m/s, and 2 m/s, and six flights are performed for data collection at each speed. Then the flight data are analyzed to verify the speed selection result, as well as the robustness of the autonomous flight system.
The third set of experiments is conducted in two vent pipes with circular cross-sections of around 0.7 m in diameter, as shown in Fig. \ref{fig:vent_case_1} and \ref{fig:vent}. These vent pipes that are commonly used in factories are adopted to validate the proposed system at the selected speed in more realistic scenarios. Comparison experiments with a SOTA motion planning method\cite{zhou2019robust} and manual flights of an experienced pilot using a commercial FPV drone shown in Fig. \ref{fig:drone_fpv} are also performed.

\begin{figure}[t]
\begin{center}
\subfigure[\label{fig:tunnel_0_viz} The visualization of the straight tunnel case shown in Fig. \ref{fig:tunnel_0}.]
{\includegraphics[width=0.47\columnwidth]{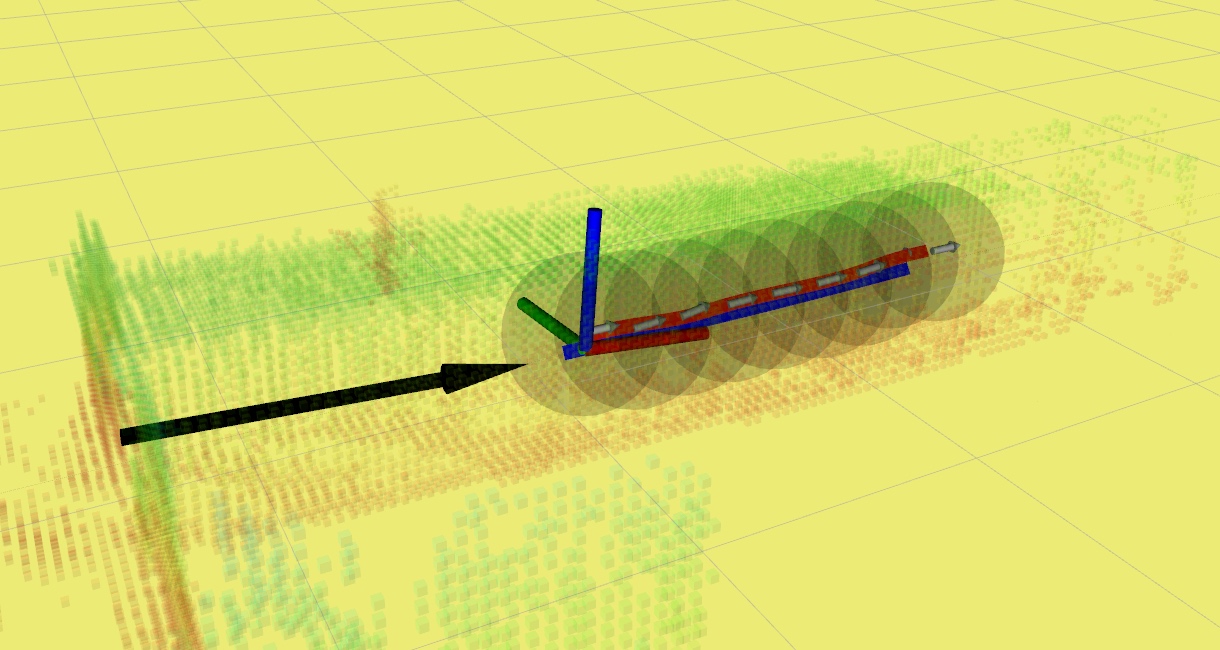}} 
\subfigure[\label{fig:tunnel_3_viz} The visualization of the curved tunnel case 1 shown in Fig. \ref{fig:tunnel_3}.]
{\includegraphics[width=0.47\columnwidth]{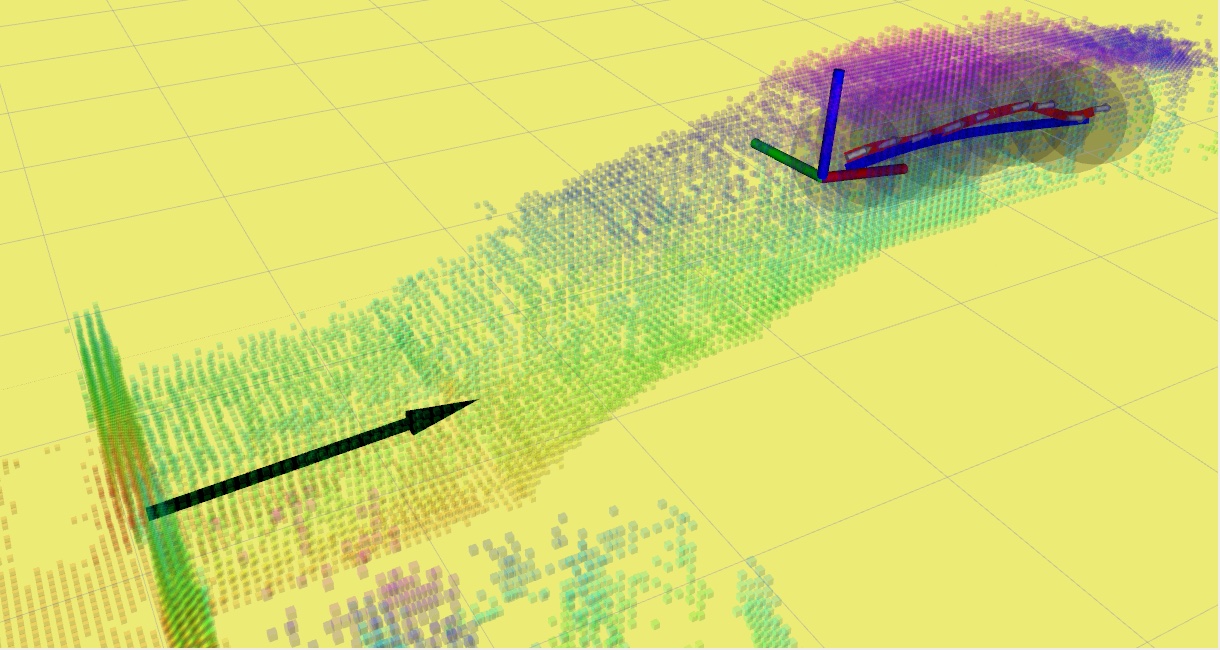}} 
\subfigure[\label{fig:tunnel_1_viz} The visualization of the curved tunnel case 2 shown in Fig. \ref{fig:tunnel_1}.]
{\includegraphics[width=0.47\columnwidth]{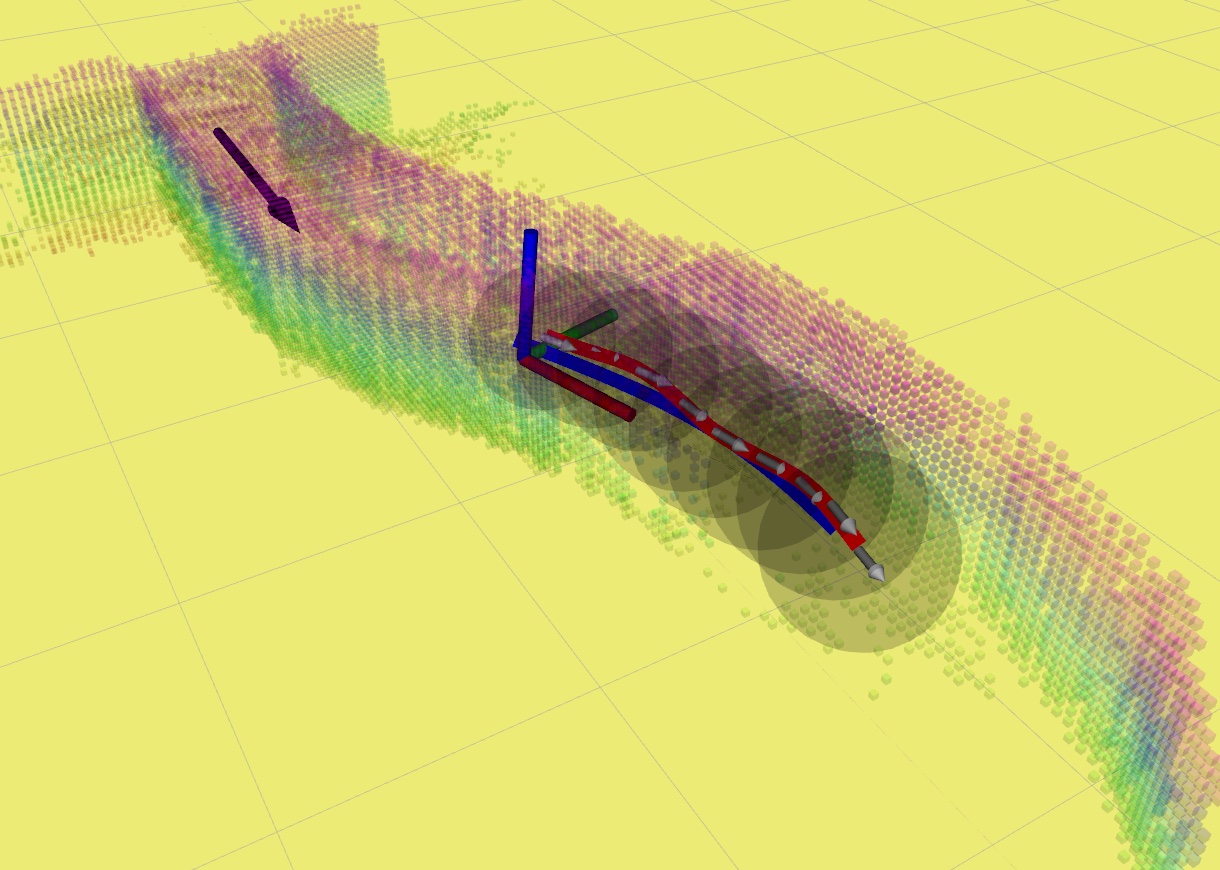}} 
\subfigure[\label{fig:tunnel_2_viz} The visualization of the curved tunnel case 3 shown in Fig. \ref{fig:tunnel_2}.]
{\includegraphics[width=0.47\columnwidth]{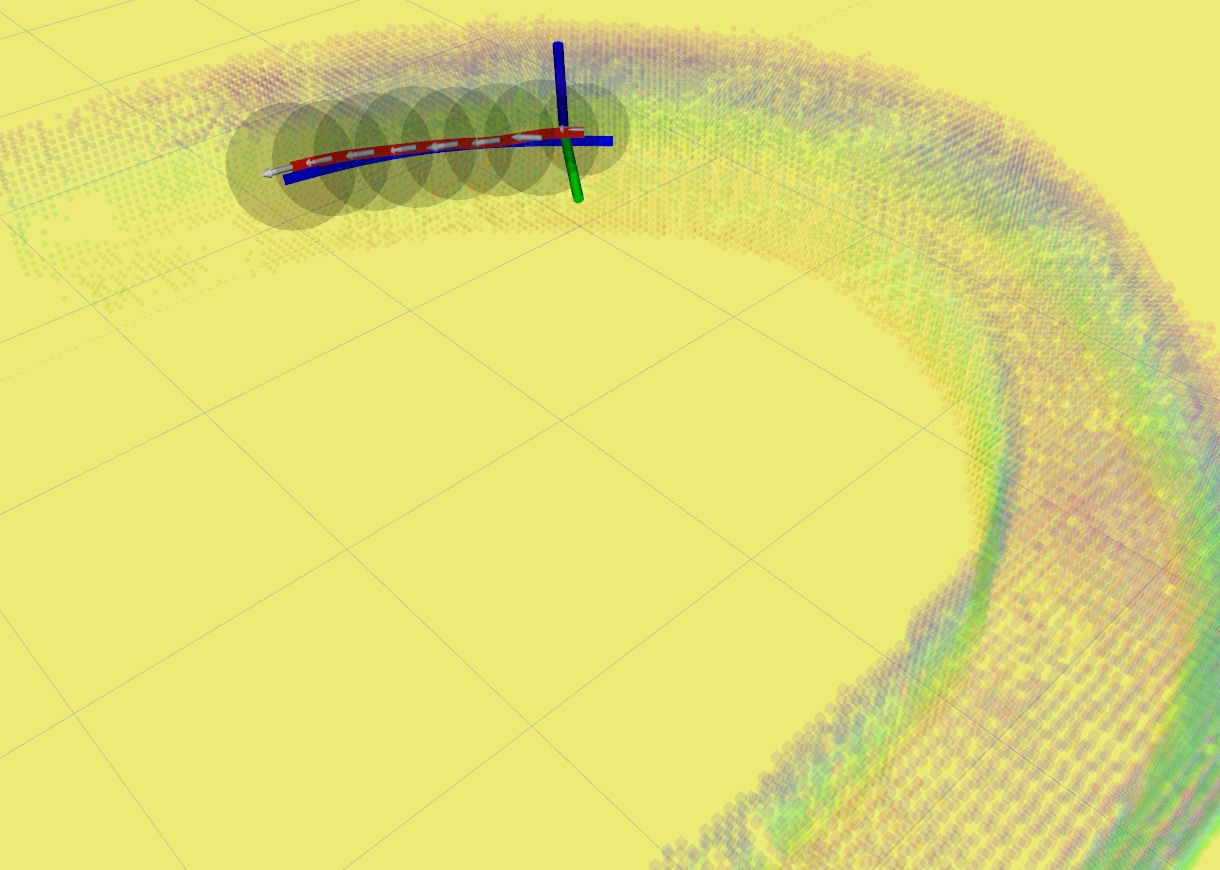}}
\subfigure[\label{fig:vent_case_1_viz} The visualization of the vent pipe case 1 shown in Fig. \ref{fig:vent_case_1}.]
{\includegraphics[width=0.47\columnwidth]{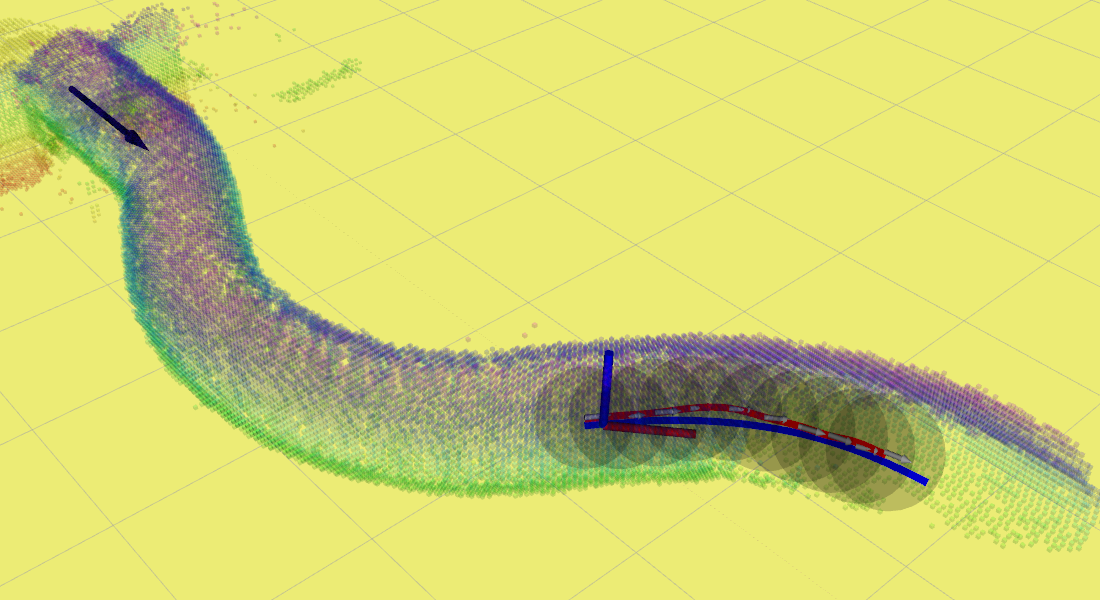}} 
\subfigure[\label{fig:vent_viz} The visualization of the vent pipe case 2 shown in Fig. \ref{fig:vent}.]
{\includegraphics[width=0.47\columnwidth]{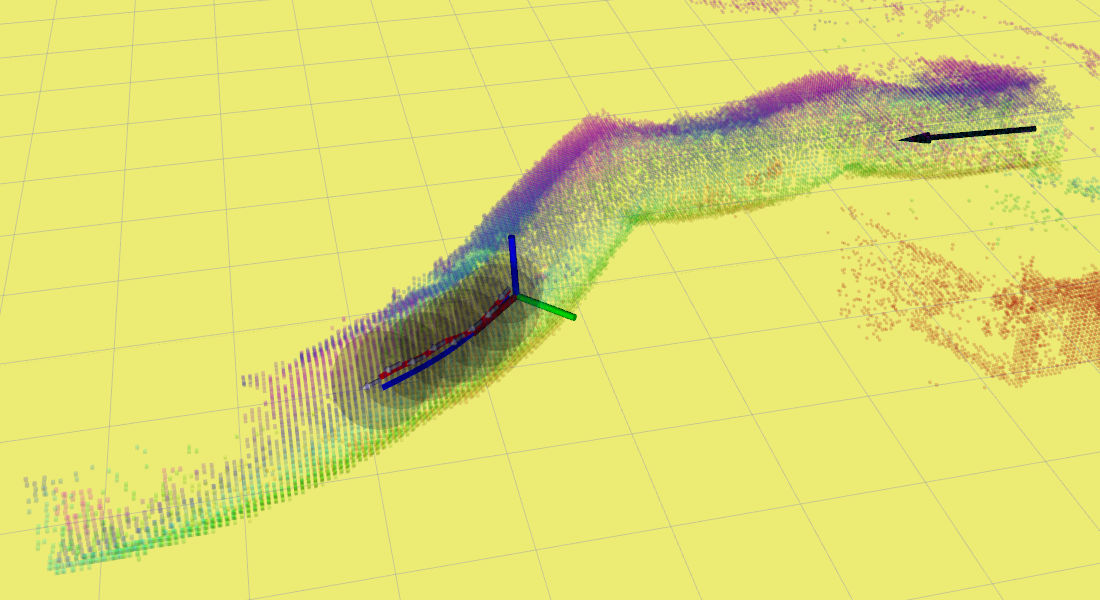}} 
\end{center}
\vspace{-0.4cm}
\caption{\label{fig:tunnel_viz} Visualization screenshots of the quadrotor flying through the narrow tunnels shown in Fig. \ref{fig:tunnel}. The color coding indicates the height and the black arrows are the estimated tunnel entrance. The axes indicate the current pose of the quadrotor. The small grey arrows indicate the searched waypoints, while the black transparent spheres indicate the spheres for the gradient descent. The red curve is the extracted tunnel center line, which is a B-spline parameterized from the waypoints, while the blue curve is the optimized trajectory for execution.}
\vspace{-1.2cm}
\end{figure}

\subsection{Straight Tunnel and Straight Line Flight Result}
\label{subsec:straight_tunnel_result}

Fig. \ref{fig:tunnel_0_viz} shows a visualization screenshot taken during the straight tunnel flights of the quadrotor using the double-phase motion planning explained in Sec. \ref{subsec:planner}. No crashes are reported during the 250 flights mentioned in Sec. \ref{subsec:exp_setup}, validating the robustness of the proposed motion planning algorithm together with the integrated system. 


\begin{figure}[t]
\begin{center}         
\subfigure[\label{fig:straight_rmse_x} The longitudinal RMSE.]
{\includegraphics[width=0.48\columnwidth]{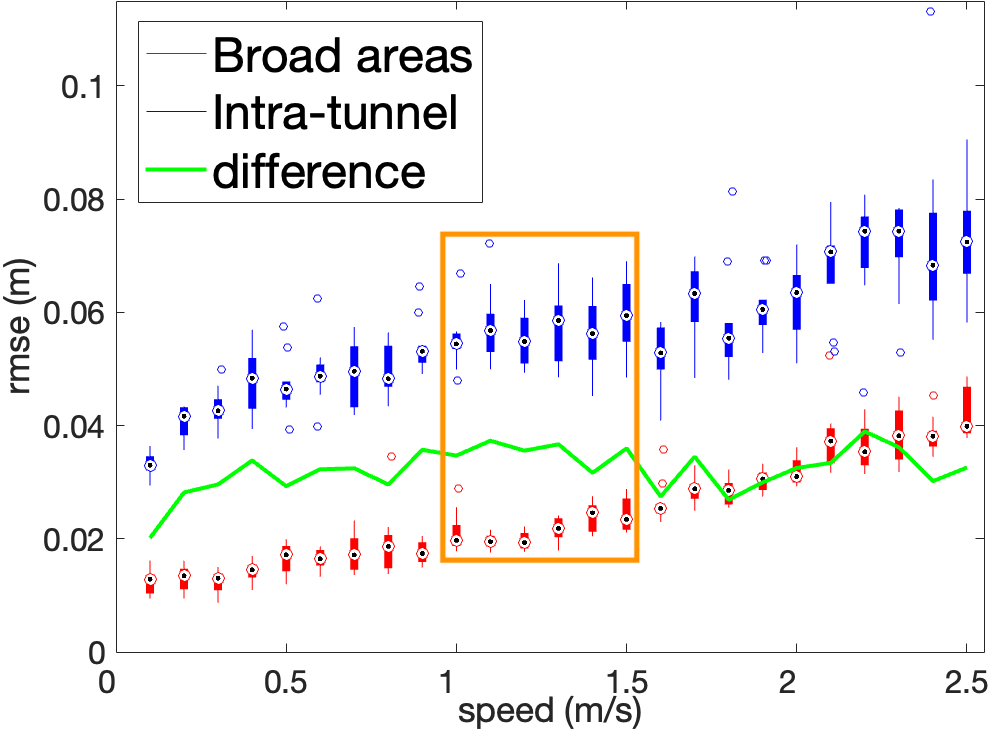}}
\subfigure[\label{fig:straight_rmse_y} The lateral RMSE.]
{\includegraphics[width=0.48\columnwidth]{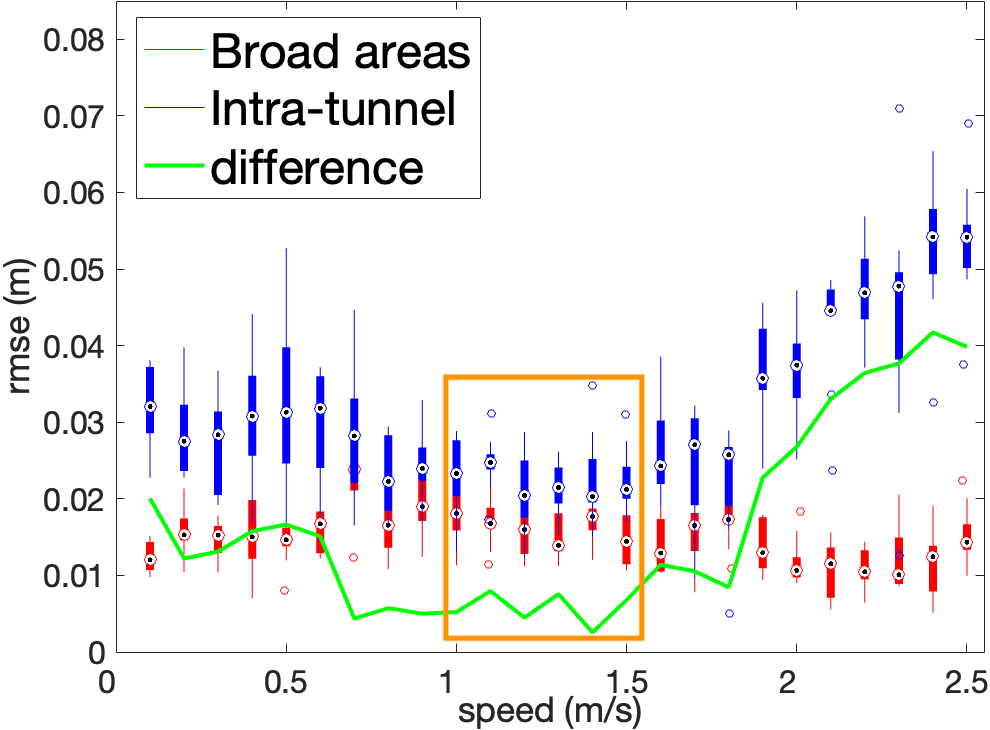}}
\subfigure[\label{fig:straight_rmse_z} The vertical RMSE.]
{\includegraphics[width=0.48\columnwidth]{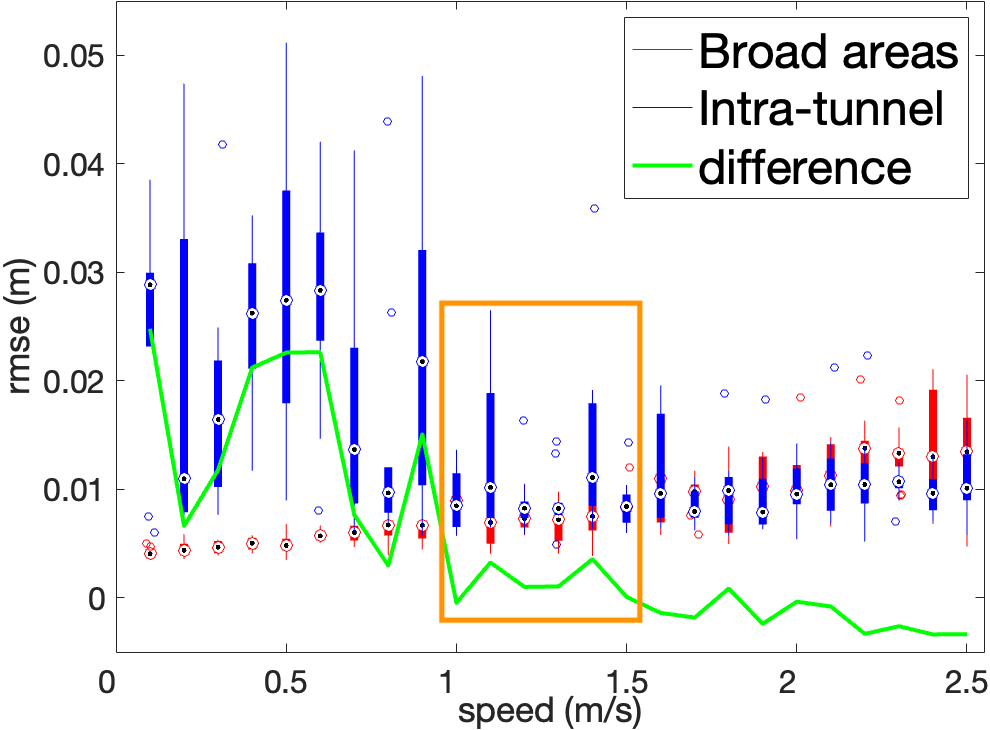}}
\subfigure[\label{fig:straight_feature} The minimum number of features tracked by the VIO system.]
{\includegraphics[width=0.48\columnwidth]{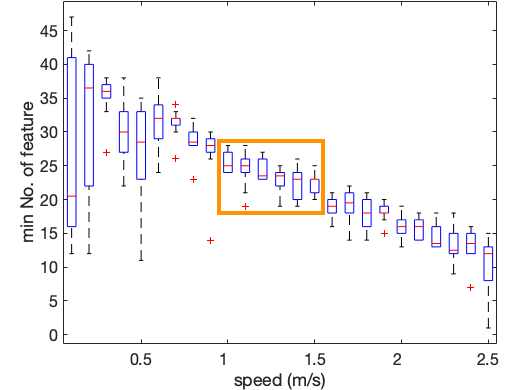}}
\end{center}
\vspace{-0.5cm}
\caption{\label{fig:straight_rmse}The box plots of the RMSE on the three directions and the minimum number of features can be tracked by the VIO system during the flights in the straight tunnel shown in Fig. \ref{fig:tunnel_0} and the comparison in position errors with the straight-line flights in broad areas outside the tunnel. The green lines indicate the error differences. The data in the optimal speed range are framed in orange rectangles.}
\vspace{-0.7cm}
\end{figure}


As stated in Sec. \ref{sec:introduction}, one of the outcomes we are interested in is the controller performance against the flight speeds. The box plots of the RMSEs of the positions during the 250 straight tunnel flights and the 250 straight line flights in broad areas are shown in Fig. \ref{fig:straight_rmse}. The plots of the error differences clearly indicate the effect on position control brought by the tunnel. In general, the position errors inside the tunnel are larger than the errors in broad areas, especially in the longitudinal direction, and the error difference in this direction indicates the longitudinal disturbance brought by the tunnel is almost constant. However, it can be observed that in the vertical direction, when the flight speed is greater than 1 m/s, the error difference almost drops to 0, i.e., the errors inside the tunnel are comparable to the errors in broad areas, indicating mitigation of the effects of the airflow disturbances compared with the scenarios with speeds of less than 1 m/s. Similar outcomes in the lateral direction can also be observed within 1.5 m/s. However, as the speed further increases, the errors increase dramatically in the lateral direction. During the experiments, when the speed rises above 2.0 m/s, it can be observed that the quadrotor diverts in most of the flights. Concurrently, as indicated in Table \ref{tab:fail_rate}, the failure in which the quadrotor makes contact with the tunnel wall occurs more and more frequently with the increment in speed, bringing about potential safety issues, despite successful traversals. In broader areas, these errors can be easily corrected by the controller. However, at a high speed in the narrow tunnel, the control bandwidth limit can be easily reached due to the unsteady flow. Meanwhile, on account of the near-wall effect, the quadrotor is more likely to be attracted to the wall as it approaches it, further enhancing the effect, and thus producing positive feedback. This eventually induces large diversions and error differences, and even collisions. Furthermore, although the longitudinal errors have no correlation to safety, the steady errors from 1 m/s to 1.5 m/s also indicate stable control performance, demonstrating the desirability of this speed range for the controller.

\begin{table}[t]
\begin{center}
\caption{Fail Rate in Straight Tunnel Flights}
\label{tab:fail_rate}
\begin{tabular}{|c|c|c|c|c|c|c|}
\hline
Speed $(m/s)$ & $\leq$ 2.0 & 2.1 & 2.2 & 2.3 & 2.4 & 2.5 
\\ 
\hline
Fail Rate & 0\% & 10\% & 30\% & 40\% & 30\% & 60\% 
\\
\hline
\end{tabular}
\end{center}
\vspace{-1.0cm}
\end{table}

The other critical result is the minimum number of features tracked by the VIO system during the tunnel flights, as shown in the box plot in Fig. \ref{fig:straight_feature}. It can be observed that the minimum number of tracked features generally decreases with the increase in the flight speed, which accords with common sense. Nonetheless, during flights at slower speeds, typically under 0.5 m/s, the variation of the data is generally larger than it is at higher speeds and occasionally the number of tracked features is even lower than that at flights at higher speeds. This phenomenon is possibly attributed to the heavy shaking motion caused by the severer flow disturbance effect at low speeds. It is also observed that within the speed range of 1 m/s to 1.5 m/s, the variance in the number of tracked features is small and those numbers are usually large enough, namely, larger than 20, which also demonstrates the practicability of this speed range.

\subsection{Curved Tunnel Result}
\label{subsec:curved_tunnel_result}



Visualization screenshots during the three curved tunnel cases are shown in Fig. \ref{fig:tunnel_3_viz}, \ref{fig:tunnel_1_viz}, and \ref{fig:tunnel_2_viz}. During the 90 flights, no failures are reported, further proving the robustness of the motion planning algorithm and the autonomous system in curved tunnels.

\begin{figure}[t]
\begin{center}         
\subfigure[\label{fig:curved_rmse_x} The longitudinal RMSE.]
{\includegraphics[width=0.48\columnwidth]{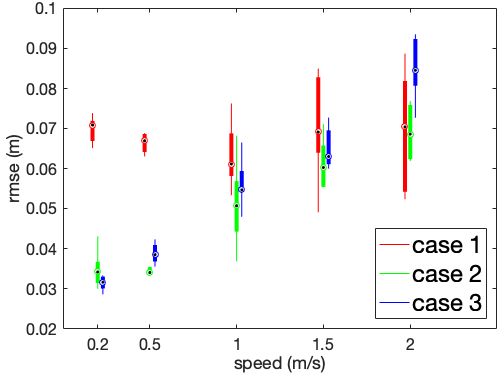}}
\subfigure[\label{fig:curved_rmse_y} The lateral RMSE.]
{\includegraphics[width=0.48\columnwidth]{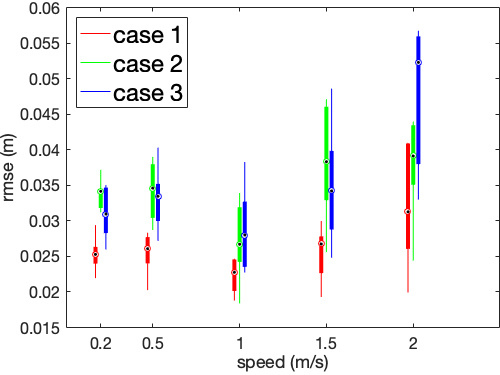}}
\subfigure[\label{fig:curved_rmse_z} The vertical RMSE.]
{\includegraphics[width=0.48\columnwidth]{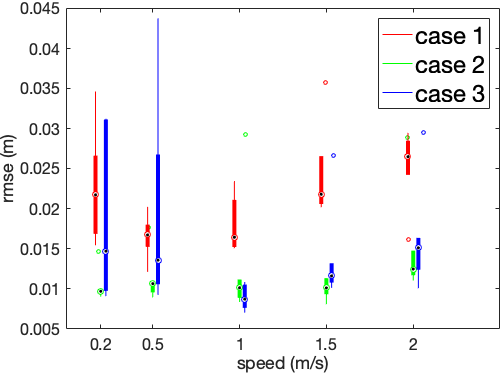}}
\subfigure[\label{fig:curved_feature} The minimum number of features tracked by the VIO system.]
{\includegraphics[width=0.48\columnwidth]{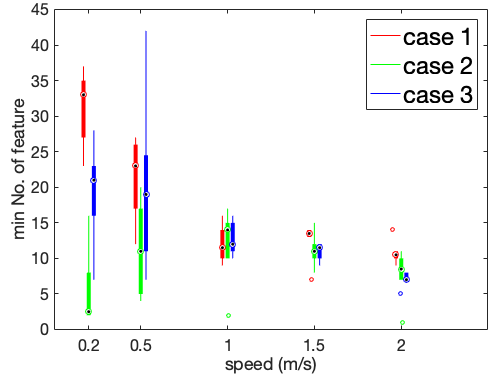}}
\end{center}
\vspace{-0.4cm}
\caption{\label{fig:curved_rmse}The box plots of the RMSE on the three directions and the minimum number of features can be tracked by the VIO system during the flights in the curved tunnels shown in Fig. \ref{fig:tunnel_3} to \ref{fig:tunnel_2}.}
\vspace{-0.7cm}
\end{figure}

The box plots in Fig. \ref{fig:curved_rmse} show the controller performance in three cases in terms of the position RMSEs against the flight speeds. It can be observed that due to the altitude changes of the tunnel in case 1, the control difficulty is increased in that direction, causing the increase in vertical error compared with the other two cases. As a consequence of their larger variation in the lateral direction, the lateral RMSEs of case 2 and 3 are greater than that of case 1. Despite the additional control difficulty brought about by the change in the lateral and vertical direction as well as the yaw, the position errors in the lateral and vertical direction reach their minimum at 1 m/s in all three cases. The errors in the longitudinal direction for case 2 and 3 increase as the speed increases, which generally aligns with the data from the previous straight tunnel flights. However, for the case 1 tunnel with vertical variations, the longitudinal RMSE reaches its minimum at 1 m/s, which may be induced by the change in the flow conditions caused by the variations, i.e., the blockage of the flow at the front or the back, turning the turbulent flow back towards the quadrotor, engendering larger disturbances at slower speeds.

The minimum number of features tracked by the VIO system is generally the same for case 1 and case 3 compared with the straight tunnel flights, as shown in Fig. \ref{fig:curved_feature}. However, for case 2, the minimum number of features is even smaller at slow speeds of 0.2 m/s and 0.5 m/s compared with that at higher speeds due to the large disturbance and yaw change in the middle position of the tunnel. Additionally, the numbers of tracked features have larger variances at slow speeds for all three cases, which also coincides with the previous results of the straight tunnel flights.

Therefore, despite the additional control difficulty, the flight data from all three curved tunnel cases demonstrate the superiority of the flight speed of 1 m/s, which is generally consistent with the proper speed range derived from the straight tunnel flight data.

\subsection{Vent Pipe Result and Comparison}
\label{subsec:vent_pipe_result}

In the vent pipe scenarios, which highly resemble the real scenes in factories, the speed of 1 m/s derived from the previous experiments is adopted in consideration of the comparable size of the vent pipe with the previous tunnels. Visualization screenshots during the two cases are shown in Fig. \ref{fig:vent_case_1_viz} and \ref{fig:vent_viz}. Despite the shape difference, the autonomous flight system traverses the pipes smoothly without any collisions. We also compare our method with a SOTA motion planning method\cite{zhou2019robust} using the same vent pipe shown in Fig. \ref{fig:vent_case_1}, and the same hardware system. However, instead of smooth flights through the pipe performed by the proposed system, the quadrotor can never fly through the pipe and even crash at the entrance using the SOTA method. Additionally, we also invite an experienced pilot to try to fly a commercial FPV drone, DJI FPV drone, which has a comparable size with our proposed quadrotor platform, through the pipes. Even with the equipment including the goggles and the remote controller, as well as the industry level drone, the pilot crashes the drone in both pipes. The comparisons in the realistic environment further prove the validity of the proposed method as well as the value and robustness of the proposed autonomous tunnel flight system.

\section{Extendability of the System}
\label{sec:extendability}
In the experiment and result, the system is proved to be adaptive and robust. It can be easily predicted that tunnels with dimensions larger than 0.6 m are easier to be traversed due to the weaker ego airflow disturbances inside\cite{powers2013influence,conyers2019empirical,wang2021estimation}. Hence, the proposed system can still be functional in wider tunnels. Additionally, the experiments have also proved the practicability in the two majority shapes of tunnels, i.e. square shape and circular shape, meaning that the system can be extended to traverse a vast number of tunnels. For another multirotor with a different size or configuration, the motion planning method is still valid, while the workflow for speed selection mentioned in Sec. \ref{subsubsec:speed_workflow} may need to be repeated for better performance.

\section{Conclusion}
\label{sec:conclusion}

In this letter, we propose an autonomous narrow tunnel flight system. Firstly, we develop a robust double-phase motion planning framework for narrow tunnels, which consists of gradient-based tunnel center line extraction and trajectory optimization. Then, the planner together with state estimation, perception, and control modules, is integrated onto a customized quadrotor platform equipped with illuminations to form a complete autonomous tunnel flight system. CFD analyses are conducted to verify the intuition regarding speed and disturbance, and extensive straight tunnel flight experiments are performed to obtain a practical flight speed range according to the control and the feature tracking data. Moreover, flights through multiple curved tunnels along with comparison experiments in vent pipes are also conducted to verify the practicability of the proposed autonomous system in more complex and realistic scenarios as well as prove the robustness of the integrated system. In this work, we assume a constant flight speed through a narrow tunnel with constant dimensions. In the future, we plan to extend the framework to flights of varying speeds in tunnels with varying dimensions by conducting further experiments and analyses.

\newlength{\bibitemsep}\setlength{\bibitemsep}{.0238\baselineskip}
\newlength{\bibparskip}\setlength{\bibparskip}{0pt}
\let\oldthebibliography\thebibliography
\renewcommand\thebibliography[1]{%
  \oldthebibliography{#1}%
  \setlength{\parskip}{\bibitemsep}%
  \setlength{\itemsep}{\bibparskip}%
}

\bibliography{ICRA2022_Luqi} 
\end{document}